\documentclass[sigconf,nonacm]{acmart}
\usepackage{adjustbox}
\usepackage[normalem]{ulem}

\usepackage{amssymb}
\usepackage{amsmath,bm}
\usepackage{pifont}
\usepackage{graphicx}
\usepackage{url}
\usepackage{xspace}
\usepackage{hyperref}
\usepackage{xcolor}
\usepackage{booktabs} 
\usepackage{makecell}
\usepackage{paralist, tabularx}
\usepackage{algorithm}
\usepackage{algorithmic}
\usepackage{caption}
\usepackage{subcaption}
\usepackage[T1]{fontenc}
\usepackage{multirow,multicol}
\usepackage{wrapfig}
\usepackage{balance}
\usepackage{bold-extra}
\useunder{\uline}{\ul}{}

\newtheorem{problem}{Problem}

\newcommand{\hide}[1]{}

\newcommand{\method}{{\sc BeneFitter}\xspace}{}

\newcommand{\tick}{\ding{51}}

\newcommand{\cbit}{\begin{compactitem}}
	\newcommand{\ceit}{\end{compactitem}}
\newcommand{\cben}{\begin{compactenum}}
	\newcommand{\ceen}{\end{compactenum}}

\newcommand{\bal}{\begin{align}}
\newcommand{\ean}{\end{align}}
\newcommand{\bit}{\begin{itemize}}
	\newcommand{\eit}{\end{itemize}}
\newcommand{\ben}{\begin{enumerate}}
	\newcommand{\een}{\end{enumerate}}
\newcommand{\beq}{\begin{equation}}
\newcommand{\eeq}{\end{equation}}

\newcommand{\etal}{\textit{et al.}\xspace}

\newcommand{\bX}{\mathbf{X}}

\newcommand{\bh}{\mathbf{h}}

\newcommand{\nvar}{d}
\newcommand{\lab}{l}
\newcommand{\len}{L}
\newcommand{\real}{\mathbb{R}}
\newcommand{\timeseries}{time-series\xspace}
\newcommand{\nclass}{C}
\newcommand{\nentities}{n}
\newcommand{\mcost}{M}
\newcommand{\save}{$s$\xspace}
\newcommand{\benefit}{{\tt benefit}\xspace}

\newcommand{\blue}[1]{\colorbox{blue!40}{#1}}
\newcommand{\gray}[1]{\colorbox{blue!25}{#1}}
\newcommand{\xmark}{\ding{55}}%

\begin{document}

\title{Benefit-aware Early Prediction of Health Outcomes on Multivariate EEG  Time Series}

\author{Shubhranshu Shekhar}

\affiliation{%
  \institution{Heinz College \& Machine Learning Dept.,  Carnegie Mellon University}
}

\author{Dhivya Eswaran}
\authornote{Now at Amazon Inc.}
\affiliation{%
  \institution{Department of Computer Science, Carnegie Mellon University}}

\author{Bryan Hooi}
\authornote{Work done while at Carnegie Mellon University}
\affiliation{%
  \institution{School of Computing, \\National University of Singapore}
}

\author{Jonathan Elmer}
\affiliation{%
 \institution{Departments of Emergency Medicine, Critical Care Medicine and Neurology, University of Pittsburgh}
}

\author{Christos Faloutsos}
\affiliation{%
  \institution{Department of Computer Science, Carnegie Mellon University}
}

\author{Leman Akoglu}
\affiliation{%
	\institution{Heinz College of Info. Sys., \\ Carnegie Mellon University}
}

\begin{abstract}

Given a cardiac-arrest patient being monitored in the ICU (intensive care unit) for brain activity, how can we predict their health outcomes as early as possible? 
Early decision-making is critical 
in many applications, 
e.g. monitoring patients
may assist in early intervention and improved care. On the other hand, 
early prediction on EEG data 
poses several challenges: (i) earliness-accuracy trade-off;  observing more data often increases accuracy but sacrifices earliness, (ii) large-scale (for training) and streaming (online decision-making) data processing,
and (iii) multi-variate (due to multiple electrodes) and multi-length (due to varying length of stay of patients) time series. Motivated by this real-world application, we present \method that
infuses the incurred savings from an early prediction as well as the cost from misclassification into a unified domain-specific target called \benefit. Unifying these two quantities allows us to directly estimate a \textit{single} target (i.e. {\tt benefit}), and importantly, 
dictates exactly \textit{when} to output a prediction: when {\tt benefit} estimate becomes positive. \method 
 (a) is {\em{efficient and fast}}, with training time linear in the number of input sequences, and can operate in real-time for decision-making,
(b) can handle \textit{multi-variate and variable-length} time-series, suitable for patient data, and 
(c) is \textit{effective}, 
providing up to 
$2\times$ time-savings 
with
equal or better accuracy as compared to competitors.
\end{abstract}

\maketitle
\pagestyle{plain}

\section{Introduction}
\label{sec:intro}



Early decision making is critical in a variety of application domains. In medicine, earliness in prediction of health outcomes for patients in ICU allows the hospital to redistribute their resources (e.g., ICU bed-time, physician time, etc.) to in-need patients, and potentially achieve better health outcomes overall within the same amount of time. Of course, another critical factor in play is the accuracy of such predictions. Hastily but incorrectly predicting unfavorable health outcome~(e.g withdrawal of life-sustaining  therapies) could hinder equitable decision making in the ICU, and may also  expose hospitals to very costly lawsuits.



A clinician considers patient history, demographics, etc. in addition to large amounts of real-time sensor information for taking a decision. Our work is motivated by the this real-world application that would help in alleviating the information overload on clinicians and aid them in early and accurate decision making in ICU, however, the setting is quite general. 
In predictive maintenance the goal is to monitor the functioning of physical devices (e.g., workstations, industrial machines, etc.) in real-time through sensors,
and to predict potential future failures as early as possible.
Here again earliness (of prediction) allows timely maintenance that prevents catastrophic halting of such systems, while hasty false-alarms take away from the otherwise healthy lifetime of a device, by introducing premature replacements that can be very costly. 

\begin{figure}
	\centering
	\includegraphics[width=0.8\linewidth]{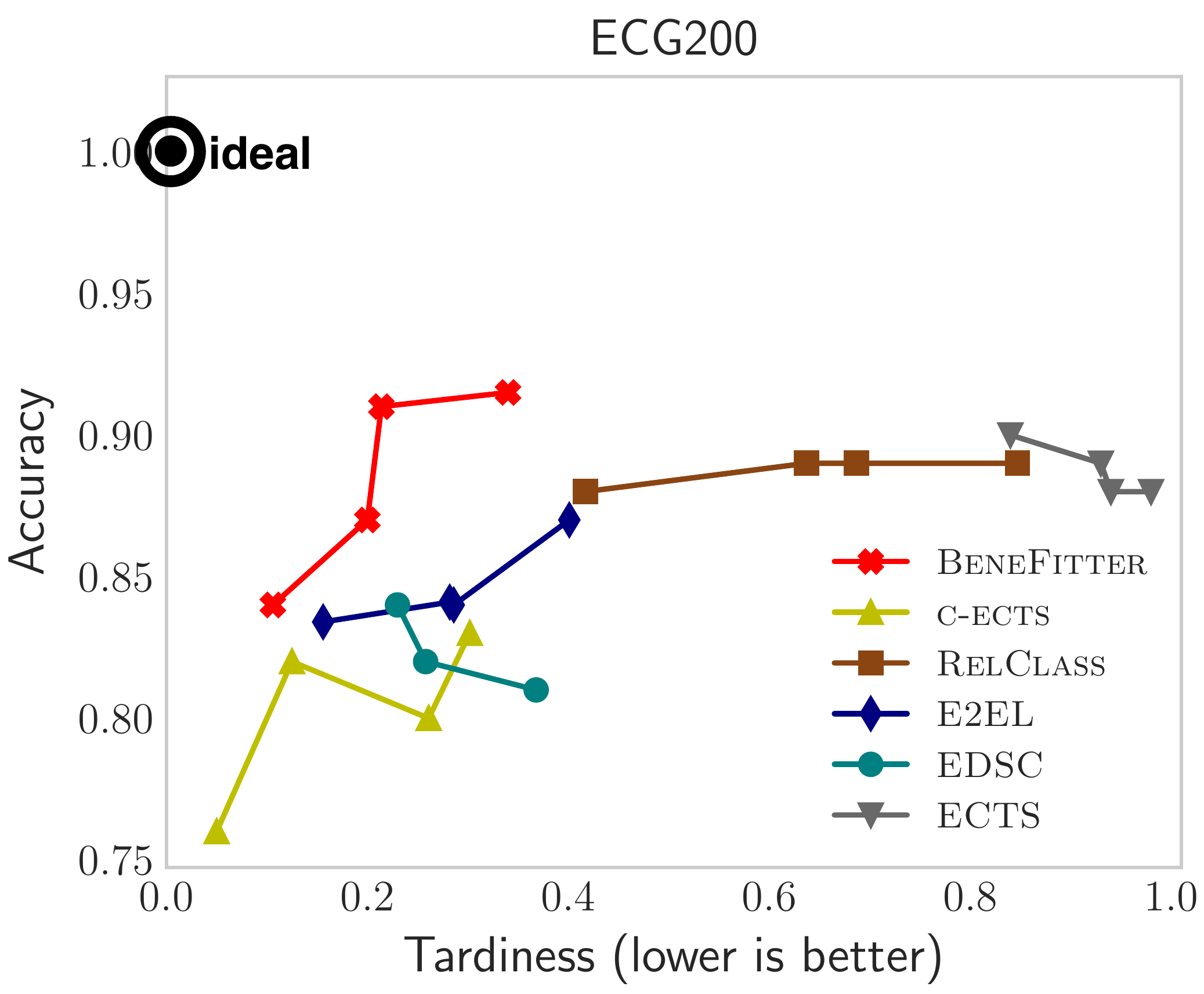}
	 \vspace*{-0.1in}
	\caption{{\bf \method wins}: Note that \method (in red) is on the Pareto front~\cite{lotov2013interactive} of accuracy-vs.-tardiness trade-off on ECG dataset. Each point represents evaluation of a method for a setting of  hyper-parameters controlling the trade-off. \label{fig:crown} 
	}
	 \vspace{-0.2in}
\end{figure}
As suggested by these applications, the real-time prediction problem necessitates modeling of two competing goals: earliness and accuracy---competing since observing for a longer time, while cuts back from earliness, provides more information (i.e., data) that can help achieve better predictive accuracy.
To this end, we directly integrate a cost/benefit framework to our proposed solution,
\method,
toward jointly optimizing prediction accuracy and earliness. 
We do not tackle an {explicit} multi-objective optimization but rather directly model a \textit{unified} target that infuses those goals. 

Besides the earliness-accuracy trade-off, the prediction of health outcomes on electroencephalography~(EEG) recordings of ICU patients brings additional challenges. A large number~(107) of EEG signal measurements are collected from multiple electrodes constituting high dimensional multivariate time series (our data is 900 GB on disk). Moreover, the series in data can be of various lengths because patients might not survive or be discharged after varying length of stay at the ICU.
\method addresses these additional challenges such as handling ($i$) multi-variate and ($ii$) variable-length signals (i.e., time series), ($iii$) space-efficient modeling, ($iv$) scalable training, and ($v$) constant-time prediction.

We summarize our contributions as follows. 

\begin{itemize}
	\setlength{\leftmargin}{0em}
	\item {\bf Novel, cost-aware problem formulation}: We propose \method, which infuses the incurred \textit{savings/gains} $S(t)$ from an early prediction at time $t$, as well as
	the \textit{cost} $M$ from each misclassification into a unified target called {\tt benefit} $=S(t)-M$. Unifying these two quantities allows us to directly estimate a \textit{single} target, i.e., {\tt benefit}, and importantly 
	dictates \method exactly \textit{when} to output a prediction: whenever estimated {\tt benefit} becomes positive.
	
	\item {\bf Efficiency and speed}: The training time for \method is linear in the number of input sequences, and it can operate under a streaming setting to update its decision based on incoming observations. Unlike existing work that train a collection of prediction models for each $t=1, 2, \ldots$~\cite{dachraoui2015early, tavenard2016cost, mori2017early}, \method employs a \textit{single} model for each possible outcome, resulting in much greater space-efficiency.

	\item {\bf Multi-variate and multi-length time-series}:
	Due to hundreds of measurements from EEG signals collected from patients with variable length stays at the ICU, 
	\method employs models that are designed to handle multiple time sequences, of varying length, which is a more general setting.

	\item {\bf Effectiveness on real-world data}: 
We apply \method on real-world (a) multi-variate health care data (our main motivating application for this work is predicting survival/death of cardiac-arrest patients based on their EEG measurements at the ICU),
and (b) other 11 benchmark datasets pertaining to various early prediction tasks. On ICU application, \method can make decisions with up to $2\times$ time-savings as compared to competitors while achieving equal or better performance on accuracy metrics. Similarly, on benchmark datasets, \method provides the best spectrum for trading-off accuracy and earliness (e.g. see Figure~\ref{fig:crown}).


\end{itemize}

\noindent
{\bf Reproducibility.~} We share all source code and public datasets at 
{{\url{https://bit.ly/3n4wL0N}}}. Our EEG dataset involves real ICU patients and is under NDA with our collaborating healthcare institution.

\noindent {\bf Suitability for Real-world Use Cases}
\cbit
\item Business problem: Customizable, cost-aware formulation. The {\tt benefit} allows domain experts to explicitly control for $S(t)$ and $M$, thereby, enabling them to incorporate into the predictions the knowledge gathered through experience or driven by the application domain. 
\item Usability: Early prediction of health outcomes to aid physicians 
in their decisions.
\ceit

\section{Data and Problem Setting}
\label{sec:motivation}
\subsection{Data Description}
\label{subsec:data}

Our use case data are
obtained from $725$ comatose patients who are resuscitated from cardiac arrest and underwent post-arrest electroencephalography~(EEG) monitoring at a single academic medical center between years $2010$--$2018$.

The raw EEG data are recorded at $256$ Hz from $22$ scalp electrodes; $11$ electrodes in each hemisphere of the brain placed according to 10--20 International System of Electrode Placement.\footnote{\url{https://en.wikipedia.org/wiki/10-20\_system\_(EEG)}} The raw data is then used to collect quantitative EEG~(qEEG) features at an interval of ten seconds that amounts to about $900$ GB of disk space for $725$ patients. For our experiments, we selected $107$ qEEG signals that physicians find informative from the electrode measurements corresponding to different brain regions. The $107$-dimensional  qEEG measurements from different electrodes on both left and right hemisphere, including the amplitude-integrated EEG (aEEG), burst suppression ratio (SR), asymmetry, and rhythmicity, 
form our multivariate time-series for analysis. We also record qEEG for each hemisphere as average of qEEG features from $11$ electrodes on the given hemisphere. 

As part of preprocessing, we normalize the qEEG features in a range $[0, 1]$. The EEG data contains artifacts caused due to variety of informative (e.g. the patient wakes up) or arbitrary (e.g. device unplugged/unavailability of devices) reason. This results in missing values, abnormally high or zero measurements. We filter out the zero measurements, typically, appearing towards the end of each sequence as well as abnormally high signal values at the beginning of each time series from the patient records. The zero measurements towards the end appear because of the disconnection. Similarly, abnormally high readings at the start appear when a patient is being plugged for measurements. The missing values are imputed through linear interpolation.

In this dataset, 225 patients ($\approx31\%$) out of total 725 patients survived i.e. woke up from coma. Since the length of stay in ICU depends on each individual patient, the dataset contains EEG records of length 24--96 hours. To extensively evaluate our proposed approach, we create $3$ versions of the dataset by median sampling~\cite{justusson1981median} the sequences at one hour, 30 minutes and 10 minutes intervals (as summarized in \S\ref{sec:exp}, Table~\ref{tab:datadesc}).

\subsection{Notation}

A multi-variate \timeseries dataset is denoted as $\mathcal{X} = \{(\bX_i, \lab_i)\}_{i=1}^{\nentities}$, consisting of observations and labels for $\nentities$ instances. Each instance $i$ has a label $\lab_i\in\{1,\ldots\nclass\}$ where $\nclass$ is the number of labels or classes.\footnote{We use the terms label and class interchangeably throughout the paper.} For example, each possible health outcome at the ICU is depicted by a class label as $survival$ or  $death$.
 The sequence of observations is given by $\bX_i = \{\bX_{i1}, \ldots, \bX_{it}, \ldots, \bX_{iL_i}\}$ for $\len_i$ equi-distant time ticks. Here, $\len_i$ is the length of \timeseries $i$ and varies from one instance to another in the general case. 
It is noteworthy that our proposed  \method can effectively handle variable-length series in a dataset, whereas most existing early prediction techniques are limited to fixed length \timeseries, where $L_i=L$ for all $i\in [n]$.
Each observation $\bX_{it} \in \real^{d}$ is a vector of $\nvar$ real-valued measurements, where $\nvar$ is the number of variables or signals. 
We denote $\bX_i$'s observations  from the start until time tick $t$ by $\bX_{i[1:t]}$.



\subsection{Problem Statement}
Early classification of time series seeks to generate a prediction for input sequence $\bX$ based on $\bX_{[1:t]}$ such that $t$ is small and $\bX_{[1:t]}$ contains enough information for an accurate prediction.
Formally,

\begin{problem}[Early classification]
Given a set of labeled multivariate time series $\mathcal{X} = \{(\bX_i, \lab_i)\}_{i=1}^{\nentities}$, learn a function $\mathcal{F}_\theta(\cdot)$ which assigns label $\hat{l}$ to a given time series $\bX_{[1:t]}$ i.e.   $\mathcal{F}_\theta(\bX_{[1:t]}) \mapsto \hat{l}$ such that $t$ is small.
\end{problem}

\noindent{\bfseries Challenges} 
The challenges in early classification are two-fold: domain-specific and task-specific, discussed as follows.

 \textbullet $\;$ {\em Domain-specific:} Data preprocessing is non-trivial since raw EEG data includes various biological and environmental artifacts. 
Observations arrive incrementally across multiple signals where the characteristics that are indicative of class labels  may occur at different times across signals which makes it difficult to find a  decision time to output a label. Moreover, each time series instance can be of different length due to 
varying length of stay of patients at the ICU which requires careful handling.

 \textbullet $\;$ {\em Task-specific:} Accuracy and earliness of prediction are competing objectives (as noted above) since observing for a longer time, while cuts back from earliness, provides more signals that is likely to yield better predictive performance.\looseness=-1

In this work, we propose \method{}~(see \S\ref{sec:meth}) that addresses all the aforementioned challenges.

\section{Background and Related Work}
\label{sec:background}

The initial mention of early classification of time-series dates back to early 2000s \cite{rodriguez2001boosting,bregon2005early} where the authors consider the value in classifying prefixes of time sequences. However, it was formulated as a concrete learning problem only recently \cite{xing2008mining,xing2012early}. Xing \etal \cite{xing2008mining}~mine a set of sequential classification rules and formulate an early-prediction utility measure to select the features and rules to be used in early classification.
Later they extend their work to a nearest-neighbor based time-series classifier approach to wait until a certain level of confidence is reached before outputting a decision \cite{xing2012early}.
Parrish \etal \cite{parrish2013classifying}~delay the decision until a reliability measure indicates that the decision based on the prefix of time-series is likely to match that based on the whole time-series.
Xing \etal \cite{xing2011extracting}~advocate the use of interpretable features called shapelets~\cite{ye2009time} which have a high discriminatory power as well as occur earlier in the time-series. Ghalwash and Obradovic \cite{ghalwash2012early} extend this work to incorporate a notion of uncertainty associated with the decision.
Hatami and Chira \cite{DBLP:conf/ciel/HatamiC13}~train an ensemble of classifiers along with an agreement index between the individual classifiers such that a decision is made when the agreement index exceeds a certain threshold.
As such, none of these methods explicitly optimize for the trade-off between earliness and accuracy.
\begin{table}[]
	\caption{\label{tab:qualitative_comparison} Qualitative comparison with prior work. `?' 
		means that the respective method, even though does not exhibit the corresponding property originally, can possibly be extended to handle it.}
	\vspace{-0.1in}
	\centering
	\resizebox{0.95\columnwidth}{!}{
		\begin{tabular}{l| rrrrrr|r}
			\toprule
			
			\textbf{Property} / \textbf{Method} &  \rotatebox{90}{ECTS~\cite{xing2012early}} & \rotatebox{90}{C-ECTS~\cite{dachraoui2015early, tavenard2016cost}} &  \rotatebox{90}{EDSC~\cite{xing2011extracting}} &  \rotatebox{90}{M-EDSC~\cite{ghalwash2012early}} &  \rotatebox{90}{RelClass~\cite{parrish2013classifying}} &  \rotatebox{90}{E2EL~\cite{russwurm2019end}} & \rotatebox{90}{\method}   \\
			\midrule
			Jointly optimize earliness \& accuracy & & & & & & \tick & \tick \\
			\hline
			Distance metric agnostic  & &  & & & \tick & \tick & \tick \\
			\hline
			Multivariate  & & & &\tick & & \tick & \tick \\
			\hline
			Constant decision time & & & & & & \tick & \tick \\
			\hline
			Handles variable length series& \tick$^{?}$  & \tick$^{?}$  & \tick &\tick &  \tick$^{?}$  & \tick & \tick \\
			\hline
			Explainable model& \tick && \tick & \tick &  && \tick\\
			\hline
			Explainable hyper-parameter  &  &  & & &  &  & \tick\\
			\hline
			Cost aware  &  &\tick  & & &  & \tick  & \tick\\
			\bottomrule
	\end{tabular}}
	\vspace{-0.2in}
\end{table}

Dachraoui \etal \cite{dachraoui2015early}~propose to address this limitation and introduce an adaptive and non-myopic approach which outputs a label when the projected cost of delaying the decision until a later time is higher than the current cost of early classification. The projected cost is computed from a clustering of training data coupled with nearest neighbor matching.
Tavenard and Malinowski \cite{tavenard2016cost}~improve upon \cite{dachraoui2015early} by eliminating the need for data clustering by formulating the decision to delay or not to delay as a classification problem.
Mori \etal \cite{mori2017early}~take a two-step approach; where in the first step classifiers are learned to maximize accuracy, and in the second step, an explicit cost function based on accuracy and earliness is used to define a stopping rule for outputting a decision. Schafer and Leser~\cite{schafer2020teaser}, instead,  utilize reliability of predicted label as stopping rule for outputting a decision. 
However, these methods require a classification-only phase followed by optimizing for trade-off between earliness and accuracy.
Recently, Hartvigsen et al.~\cite{hartvigsen2019adaptive} employ recurrent neural network (RNN) based discriminator for classification paired with a reinforcement learning task to learn halting policy. The closest in spirit to our work is the recently proposed end-to-end learning framework for early classification~\cite{russwurm2019end} that employs RNNs. They use a cost function similar to \cite{mori2017early}  in a fine-tuning
framework to learn a classifier and a stopping rule based on RNN embeddings for partial sequences. 

Our proposed \method is a substantial improvement over all the above prior work on early classification of time series along a number of fronts, as summarized in Table~\ref{tab:qualitative_comparison}. \method jointly optimizes for earliness and accuracy using a cost-aware \texttt{benefit} function. It seamlessly handles multi-variate and varying-length time-series and moreover, leads to explainable early predictions, which is important in high-stakes domains like health care.

\section{\method: Proposed Method}
\label{sec:meth}



%
%
%

%


\subsection{Modeling Benefit}
How should an early prediction system trade-off accuracy vs. earliness? In many real-world settings, there is natural misclassification \textit{cost}, denoted $M$, associated with an inaccurate prediction and certain \textit{savings}, denoted $S(t)$, obtained from early decision-making. We propose to construct a single variable called \texttt{benefit} which captures the overall value (savings minus cost) of outputting a certain decision (i.e., label) at a certain time $t$, given as
\beq
\label{eq:benefit}
 {\tt benefit} =S(t)-M
\eeq
 We directly incorporate \texttt{benefit} into our model and leverage it in deciding {\em when} to output a decision; when the estimate is positive.

\subsubsection{\bf Outcome vs. Type Classification}
There are two subtly different problem settings that arise in \timeseries classification that are worth distinguishing between.

\newcommand{\favorable}{favorable\xspace}
\newcommand{\unfavorable}{unfavorable\xspace}

	
	\textbullet $\;$ 	\textit{Outcome classification:} \sloppy{Here, the labels of \timeseries encode the observed outcome \textit{at the end} of the monitoring period of each instance. Our motivating examples from predictive health care and system maintenance fall into this category. Typically, there are two outcomes: \textit{\favorable} (e.g., $survival$ or $no$-$failure$) and \textit{\unfavorable} (e.g., $death$ or $catastrophic$-$failure$); and we are interested in knowing when an \unfavorable outcome is anticipated. In such cases, predicting an early \favorable outcome does {not} incur any change in course of action, and hence does not lead to any discernible savings or costs.
	For example, in our ICU application, a model predicting $survive$ (as opposed to $death$) simply suggests to the physicians that the patient would survive {\em provided they continue with their regular procedures of treatment}. That is because $l=survive$ labels we observe in the data are {\em at the end} of the observed period \textit{only after all regular course of action have been conducted}.}
In contrast, $l=death$ instances have died {\em despite} the treatments. 

In outcome classification, predicting the \favorable class simply corresponds to the `default state' and therefore we model  {\tt benefit} and actively make predictions only for the \unfavorable class.

\textbullet $\;$  \textit{Type classification:} Here, the \timeseries labels capture the underlying process that gives rise to the sequence of observations. In other words, the class labels are \textit{prior} to the \timeseries observations. The standard \timeseries early classification benchmark datasets fall into this category. Examples include predicting the type of a bird from audio recordings or the type of a flying insect (e.g., a mosquito) from their wingbeat frequencies \cite{Batista2011}. Here, prediction of \textit{any} label for a \timeseries at a given time has an associated cost in case of misclassification (e.g., inaccurate density estimates of birds/mosquitoes) as well as potential savings for earliness (e.g., battery life of sensors).
In type classification, we separately model {\tt benefit} for each class.


\subsubsection{\bf Benefit Modeling for Outcome Classification}
Consider the 2-class problem that arises in predictive health care of ICU patients and predictive maintenance of systems. Without loss of generality, let us denote by $\lab=0$ the $survival$ class where the patient is discharged alive from the ICU at the end of their stay; 
and let $\lab=1$ denote the $death$ class where the patient is deceased.

As discussed previously,  $\lab=0$ corresponds to the `default state' in which regular operations are continued. Therefore, predicting $survival$ would not incur any time savings or misclassification cost.  
In contrast, predicting $death$ would suggest the clinician to intervene to optimize quality of life for the patient.
In case of an accurate prediction, say at time $t$, earliness would bring savings (e.g., ICU bed-time), denoted $S(t)$.
Here we use a linear function of time for savings on accurately predicting $death$ for a patient $i$ at time $t$, specifically\looseness=-1 
\beq
S(t) = (L_i - t)s
\eeq
where \save denotes the value of savings  
per unit time.\footnote{Note that \method is flexible enough to accommodate any other function of time, including nonlinear ones, as the savings function $S(t)$.}
On the other hand, an inaccurate $death$ flag at $t$, while comes with the same savings, would also incur a misclassification cost \mcost{} (e.g., a lawsuit).

All in all, the benefit model for the ICU scenario is given as in Table \ref{tab:icu}, reflecting the relative savings minus the misclassification cost for each decision at time $t$ on \timeseries instance $i$.
As we will detail later in \S\ref{ssec:predicting}, the main idea behind \method is to learn a single \textit{regressor} model for the $death$ class, estimating the corresponding {\tt benefit} at each time tick $t$.
\begin{table}[]
		\caption{Benefit model for ICU outcome prediction. 
	\label{tab:icu}	}
		\vspace{-0.1in}
\centering
\begin{tabular}{@{}cc|cc@{}}
	\multicolumn{1}{c}{} &\multicolumn{1}{c}{} &\multicolumn{2}{c}{Predicted $\widehat{l}_i$} \\
		\multirow{3}{*}{\rotatebox[origin=tr]{90}{Actual $l_i\;\;\;$}}  & \multicolumn{1}{c|}{} & 	\multicolumn{1}{c}{$survival$} & 	\multicolumn{1}{c}{$death$}\\ 
	\cline{2-4}
  & $survival$  & 0 & $(L_i - t)s-M$   \\[1.5ex]
	& $death$  & 0   & $(L_i - t)s$ \\ 
	\cline{2-4}
\end{tabular} 
\vspace{-0.1in}
\end{table}

\noindent{\bf Specifying $s$ and $M$.~} Here, we make the following two remarks. First, unit-time savings $s$ and misclassification cost $M$ are value estimates that are dictated by the specific application. For our ICU case, for example, we could use $s=\$4,000$ value per unit ICU time, and $M=\$500,000$ expected cost per  lawsuit. Note that $s$ and $M$ are domain-specific explainable quantities. Second, the benefit model is most likely to differ from application to application.
For example in predictive system maintenance, savings and cost would have different semantics, assuming that early prediction of failure implies a renewal of all equipment.
In that case, an early and accurate failure prediction would incur savings from costs of a complete system halt, but also loss of equipment lifetime value due to early replacement plus the replacement costs. On the other hand, early but inaccurate prediction (i.e., a false alarm) would simply incur unnecessary replacement costs plus the loss of equipment lifetime value due to early replacement.

Our goal is to set up a general prediction framework that explicitly models {\tt benefit} based on incurred savings and costs associated with individual decisions, whereas the scope of specifying those savings and costs are left to the practitioner. 
We argue that each real-world task should strive to explicitly model {\tt benefit}, where earliness and accuracy of predictions translate to real-world value.
In cases where the prediction task is isolated from its real-world use (e.g., benchmark datasets), one could set both $s=M=1$ for unit savings per unit time earliness and unit misclassification cost per incorrect decision.
In those cases where $M$ is not tied to a specific real-world value, it can be used as a ``knob'' (i.e., hyperparameter) for trading off accuracy with earliness; where, fixing $s=1$, a larger $M$ nudges \method to avoid misclassifications toward higher accuracy at the expense of delayed predictions and vice versa.

\subsubsection{\bf Benefit Modeling for Type Classification}
Compared to outcome prediction where observations give rise to the labels, in type classification problems the labels give rise to the observations. Without a default class, predictions come with associated savings and cost for each class.

\begin{table}[H]
	\caption{Benefit model for general two-class type prediction. \label{tab:twoclass}}
	\vspace{-0.3in}
	\centering
	\begin{tabular}{@{}cc|cc@{}}
		\multicolumn{1}{c}{} &\multicolumn{1}{c}{} &\multicolumn{2}{c}{Predicted $\widehat{l}_i$} \\ 
		\multirow{2}{*}{\rotatebox[origin=tr]{90}{Actual $l_i\;\;\;$}} & 
		\multicolumn{1}{c|}{} & 
		\multicolumn{1}{c}{$type$-$1$} & 
		\multicolumn{1}{c}{$type$-$2$} \\ 
		\cline{2-4}
		
		& $type$-$1$  & $(L_i - t)s$ & $(L_i - t)s-M_{12}$   \\[1.5ex]
		& $type$-$2$  & $(L_i - t)s-M_{21}$   & $(L_i - t)s$ \\ 
		\cline{2-4}
	\end{tabular} 
	\vspace{-0.1in}
\end{table}

Consider the 2-class setting of predicting an insect's type from wingbeat frequencies.
An example benefit model is illustrated in Table \ref{tab:twoclass}, $s$ capturing the value of battery-life savings per unit time and $M$ depicting the cost of misclassifying one insect as the other.
Note that in general, misclassification cost need not be symmetric among the classes.

For type classification problems, we train a total of $C$ \benefit prediction models, one for each class.
Since misclassification costs are already incorporated into \benefit,
we train each (regression) model independently which allows for full parallelism.

\subsection{Online Decision-making using Benefit}
Next we present how to employ \method in decision making in real time.
Suppose we have trained our model that produces {\tt benefit} estimates per class for a new \timeseries instance in an online fashion. 
\textit{How} and \textit{when} should we output predictions?

\begin{figure}[!ht]
	\centering
	\includegraphics[scale=0.36]{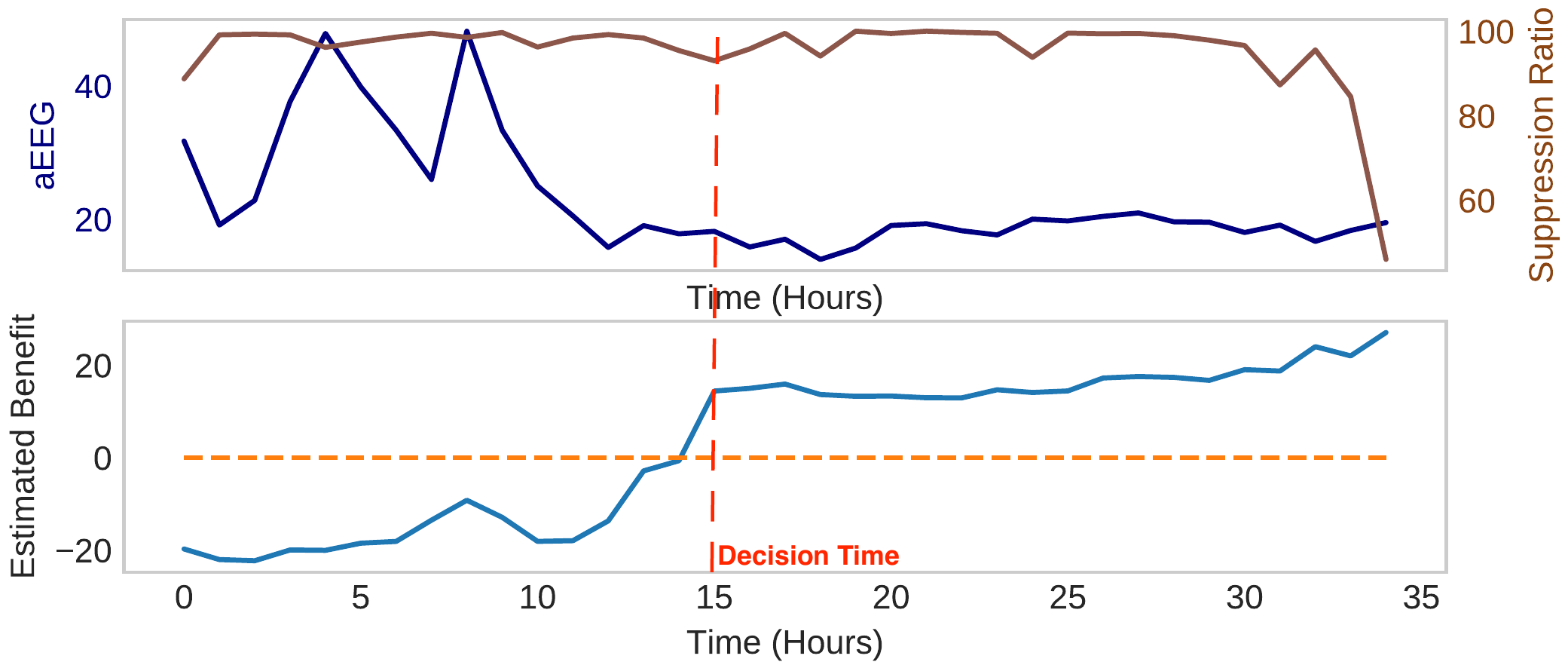}
	\vspace*{-0.2in}
	\caption{Benefit estimate over time for a patient from EEG dataset with true $\lab=1$ (i.e. $death$). We show two out of all 107 signals used by \method: amplitude of EEG (aEEG) and suppression ratio~(i.e. fraction of flat EEG epochs). \label{fig:benefit_estimate}}
	\vspace*{-0.2in}
\end{figure}

Thanks to our benefit modeling, the decision-making is quite natural and intuitive: \method makes a prediction only when the estimated {\tt benefit} becomes \textit{positive} for a certain class and outputs the label of that class as its prediction i.e. for our ICU scenario the predicted label $\hat{l}$ is given as
\[
\hat{l} =
\begin{cases}
	\text{\unfavorable}, & \text{if } {\tt benefit} > 0\\
	\text{\favorable, i.e. no action}, & \text{otherwise.}
\end{cases}
\]
 For illustration, in Fig.~\ref{fig:benefit_estimate} we show \benefit estimates over time for an input series where $t=15$ corresponds to decision time.

Note that in some cases \method may restrain from making any prediction for the entire duration $L$ of a test instance,
that is when estimated {\tt benefit} never goes above zero.
For outcome classification tasks, such a case is simply registered as default-class prediction and its prediction time is recorded as $L$. For the ICU scenario, a non-prediction is where no $death$ flag is raised, suggesting survival and regular course of action.
For type classification tasks, in contrast, a non-prediction suggests ``waiting for more data'' which, at the end of the observation period, simply implies insufficient evidence for any class.
We refer to those as un-classified test instances.
Note that \method is different from existing prediction models that always produce a prediction, where un-classified instances may be of independent interest to domain experts in the form of outliers, noisy instances, etc.

\subsection{Predicting Benefit}
\label{ssec:predicting}

For each \timeseries $i$, we aim to predict the {\tt benefit} at every time tick $t$, denoted as $b_{it}$.
Consider the outcome classification problem, where we are to train one regressor model for the non-default class, say $no-survival$.
For each training series $i$ for which $l_i=0$ (i.e., default class), \benefit of predicting $death$ at t is $b_{it}= (L_i - t)s-M$. Similarly for training series for which $l_i=1$ (i.e., $death$), $b_{it}= (L_i - t)s$. (See Table \ref{tab:icu}.)
To this end, we create training samples of the form $\big\{ (\bX_{i[1:t]}, b_{it}) \big \}_{t=1}^{L_i}$ per instance $i$.
Note that the problem becomes a regression task.
For type classification problems, we train a separate regression model per class with the corresponding $b_{it}$ values. (See Table \ref{tab:twoclass}.)


\noindent{\bfseries Model.}
We set up the task of {\tt benefit} prediction as a sequence regression problem. We require \method to ingest multi-variate and variable-length input to estimate {\tt benefit}. We investigate the use of Long Short Term Memory~(LSTM)~\cite{hochreiter1997long}, a variant of recurrent neural networks~(RNN), for the sequence (time-series) regression since their recursive formulation allows LSTMs to handle multi-variate variable-length inputs naturally. The recurrent formulation of LSTMs is useful for \method to enable real-time predictions when new observations arrive one at a time.

\noindent{\bfseries  Attention.}
 The recurrent networks usually find it hard to focus on to relevant information in long input sequences. For example, an EEG pattern in the beginning of a sequence may contain useful information about the patient's outcome, however the lossy representation of LSTM would forget it. This issue is mostly encountered in longer input sequences~\cite{luong2015effective}. The underlying idea of \emph{attention}~\cite{vaswani2017attention} is to learn a context that captures the relevant information from the parts of the sequence to help predict the target. For a sequence of length $L$, given the hidden state $\bh_t$ from LSTM and the context vector $\boldsymbol{c}$, the attention step in \method combines the information from both vectors to produce a final attention based hidden state as described below:
\begin{align}
\alpha_t &= \frac{\exp\big(\boldsymbol{c}_L \cdot\bh_t \big)}{\sum_{t}\exp\big(\boldsymbol{c}_L \cdot \bh_t\big)}; \quad
\boldsymbol{c} = \sum_{t=1}^{L} \alpha_t \bh_t\\
\bh_{\text{attn}} &= \sigma(\mathbf{W}_a [concat(\boldsymbol{c}, \boldsymbol{c}_L)])
\end{align} 
where $\boldsymbol{c}_L$ is the memory state of the cell at the last time step $L$, $\bh_t$ is the hidden state output of the LSTM at time $t$, $\bh_{\text{attn}}$ is the attention based hidden state, $\sigma(\cdot)$ is the non-linear transformation, and $\mathbf{W}_a$ is the parameter. Intuitively, the attention weights $\alpha_t$ allows the model to learn to focus on specific parts of the input sequence for the task of regression. The \benefit prediction is given by a single layer neural network such that $ \hat{b}_L = \bh_{\text{attn}} \mathbf{w} + w_0$ where $\mathbf{w}$ and $w_0$ are parameters of the linear layer.

\method is used in real life decision making, where the attention mechanism could help an expert by highlighting the relevant information that guided the model to output a decision. We present model implementation details and list of tunable parameters in \S\ref{sec:exp}.

\section{Experiments}
\label{sec:exp}
We evaluate our method through extensive experiments on a set of benchmark datasets and on a set of datasets from real-world use cases. We next provide the details of the datasets and the experimental setup, followed by results.
\begin{table}
	\caption{Summary of the datasets used in this work.\label{tab:datadesc}}
	\vspace{-0.1in}
	\resizebox{0.9\columnwidth}{!}{
		\begin{tabular}{@{}rrrrrc@{}}
			\toprule
			Dataset          & Train & Test  & Classes & Length & Dimension \\ \midrule
			EEG-ICU Hour          & 507   & 218   & 2       & 24--96  & 107     \\ 
			EEG-ICU 30 Min         & 507   & 218   & 2       & 48--192  & 107       \\ 
			EEG-ICU 10 Min         & 507   & 218   & 2       & 144--576  & 107       \\ \midrule
			ECG200           & 100   & 100   & 2       & 96     & 1         \\
			ItalyPowerDemand & 67    & 1029  & 2       & 24     & 1         \\
			GunPoint           & 50   & 150   & 2       & 150     & 1         \\
			TwoLeadECG           & 23   & 1139   & 2       & 82     & 1         \\
			Wafer            &   1000    &    6062   &    2     &   152     &    1       \\
			ECGFiveDays            &   23    &    861   &    2     &   136     &    1       \\
			MoteStrain            &   20    &    1252   &    2     &   84     &    1       \\
			Coffee & 28   & 28   & 2      & 286     & 1         \\
			Yoga              & 300    & 3000   & 2       & 426   & 1         \\
			SonyAIBO            & 20   & 601   & 2      & 70    & 1         \\
			Endomondo        & 99754 & 42751 & 2       & 450    & 2         \\
			\bottomrule
		\end{tabular}
	}
	\vspace{-0.1in}
\end{table}

\subsection{Dataset Description}
We apply \method on our EEG-ICU datasets (see \S\ref{subsec:data}), and on $11$ public benchmark datasets from diverse domains with varying dimensionality, length and scale. Table~\ref{tab:datadesc} provides a summary of the datasets used in evaluation. Note that EEG-ICU datasets are variable-length, but benchmarks often used in the literature are not. 
Detailed description of public datasets are included in  Appx. \ref{sec:supp_data_desc}.\looseness=-1

\subsection{Experimental Setup}

{\bfseries Baselines.~} We compare \method to the following five early time-series classification methods~(also see Table~\ref{tab:qualitative_comparison}):
\begin{enumerate}
	\item ECTS: Early Classification on Time Series~\cite{xing2012early} uses \emph{minimum prediction length}~(MPL) and makes predictions if the MPL of the top nearest neighbor (1-NN) is greater than the length of the test series.
	\item EDSC: Early Distinctive Shapelet Classification~\cite{xing2011extracting} extracts local shapelets for classification that are ranked based on the utility score incorporating earliness and accuracy. Multivariate extension of EDSC (M-EDSC)~\cite{ghalwash2012early} provides a utility function that can incorporate multi-dimensional series.
	\item C-ECTS: Cost-aware ECTS~\cite{dachraoui2015early, tavenard2016cost} trades-off between a misclassification cost and a cost of delaying the prediction, and estimates future expected cost at each time step to determine the optimal time instant to classify an incoming time series.
	\item RelClass:Reliable Classification~\cite{parrish2013classifying} uses a reliability measure to estimate the probability that the assigned label given incomplete data (at time step $t$) would be same as the label assigned given the complete data.
	\item E2EL: End-to-end Learning for Early Classification of Time Series~\cite{russwurm2019end} optimizes a joint cost function based on accuracy and earliness, and provides a framework to estimate a stopping probability based on the cost function.
\end{enumerate}

\subsection{Evaluation}
\label{subsec:evaluation}
We design our experiments to answer the following questions: 

\vspace{0.05in}
\noindent{\bfseries [Q1] Effectiveness:} How effective is \method at early prediction on time series compared to the baselines? What is the trade-off with respect to accuracy and earliness? How does the accuracy--earliness trade-off varies with respect to model parameters?  

\vspace{0.05in}
\noindent{\bfseries [Q2] Efficiency:} How does the running-time of \method scale w.r.t. the number of training instances? How fast is the online response time of \method? 

\vspace{0.05in}
\noindent{\bfseries [Q3] Discoveries:}  Does \method lead to interesting discoveries on real-world case studies?

\subsubsection*{\bfseries [Q1] Effectiveness}
We compare \method to baselines on (1) patient outcome prediction, the main task that inspired our {\tt benefit} formulation, (2) the activity prediction task on a web-scale dataset, as well as (3) the set of 10 two-class time-series classification datasets. The datasets for the first two tasks are multi-dimensional and variable-length that many of the baselines can not handle. Thus we compare \method with {\scshape{E2EL}} and {\scshape{M-EDSC}} baselines that can work with such time-series sequences. Comparison with {\scshape{M-EDSC}} is limited to the smaller one-hour EEG dataset since it does not scale to larger datasets. In order to compare \method to all other baselines, we conduct experiments on ten benchmark time-series datasets.\looseness=-1

\noindent{\bfseries Patient Outcome Prediction.~} We compare \method with the baseline {\scshape{E2EL}} on two competing criteria:  performance (e.g. precision, accuracy) and earliness~(tardiness -- lower is better) of the decision. We report precision, recall, F1 score, accuracy, tardiness and the total \benefit using each method when applied to the test set. EEG dataset is a high dimensional variable-length dataset for which most of the baselines are not applicable. In our experiments, we set $M/s=\{100, 200, 600\}$ as misclassification cost for each of the dataset variants -- sampled at an hour, 30 minutes, 10 minutes --  respectively based on average daily cost of ICU care and the lawsuit cost. For the baseline methods, we report the best results for the earliness-accuracy trade-off parameters. For the baseline methods, we select the best value of accuracy and earliness based on their Euclidean distance to ideal accuracy $=1$ and ideal tardiness $=0$.

Table~\ref{tab:eeg} reports the evaluation against different performance metrics. Note that predicting `default state' for a patient does not change the behavior of the system. However, predicting \emph{death} (unfavorable outcome) may suggest clinician to intervene with alternative care. In such a decision setting, it is critical for the classifier to exhibit high precision. Our results indicate that \method achieves a significantly higher precision~(according to the micro-sign test~\cite{yang1999re}) when compared to the baselines. On the other hand, a comparatively lower tardiness indicates that \method requires conspicuously less number of observations on average to output a decision (no statistical test conducted for tardiness). We also compare the total \benefit accrued for each method on the test set where \method outperforms the competition. The results are consistent across the three datasets of varying granularity from hourly sampled data to 10 minute sampled data.

\begin{table}[!ht]
	\caption{Effectiveness of \method on EEG datasets.  * indicates significance at $p$-value $\leq 0.05$ based on the micro-sign test~\cite{yang1999re} for the performance metrics. No statistical test conducted for tardiness and total \benefit. \label{tab:eeg}}
	\vspace{-0.1in}
	\resizebox{\columnwidth}{!}{
		\begin{tabular}{@{}lrcccccr@{}}
			\toprule
			&  &\rotatebox{75}{Prec.} &\rotatebox{75}{Recall} &\rotatebox{75}{F1} & \rotatebox{75}{Acc.} & \rotatebox{75}{Tardiness} &  \rotatebox{75}{\benefit}\\ \midrule
			\multirow{3}{*}{\rotatebox{0}{EEG Hour}}  & E2EL                 & 0.70                          & \textbf{0.68}                     & 0.69                         & 0.79                         & 1.0           & -2600                \\
			& M-EDSC               & 0.69                          & 0.65                       & 0.67                         & 0.78                         & \textbf{0.52}         & 2497                 \\
			& \method            & \textbf{0.80}*                          & \textbf{0.68 }                      & \textbf{0.73}*                         & \textbf{0.83}*                         & 0.64            & 2737               \\ \midrule
			\multirow{2}{*}{\rotatebox{0}{EEG 30Min}} & E2EL                 & 0.64                          & \textbf{0.67}                       & 0.65                         & 0.78                         & 1.0     &-4800                      \\
			& \method           & \textbf{0.68}*                         & 0.66                       & \textbf{0.67}*                        & \textbf{0.79}                         & \textbf{0.63 }     & 5962                    \\ \midrule
			\multirow{2}{*}{\rotatebox{0}{EEG 10Min}} & E2EL                 & 0.73                          & \textbf{0.69 }                      & 0.71                         & 0.82                         & 0.86          & -736                \\
			& \method            & \textbf{0.76}*                          & \textbf{0.69}                       & \textbf{0.72}                         & \textbf{0.83 }                        & \textbf{0.48 }        &18722                 \\ \bottomrule
	\end{tabular}}
\end{table}

For hourly sampled set, we also compare our method to multivariate {\scshape{EDSC}} baseline (for the 30 min and 10 min EEG dataset  {\scshape{M-EDSC}} does not scale ). Though {\scshape{M-EDSC}} provides better earliness trade-off compared to other two methods, the precision of the outcomes is lowest which is not desirable in this decision setup. In Table~\ref{tab:eeg}, we indicate the significant results using * that is based on the comparison between \method and {\scshape{E2EL}}.

\begin{figure*}[!ht]
	\begin{adjustbox}{minipage=\textwidth,height=0.35\textwidth, width=0.9\textwidth, }
	\centering
	\begin{subfigure}{0.32\textwidth}
		\centering
		\includegraphics[scale=0.27]{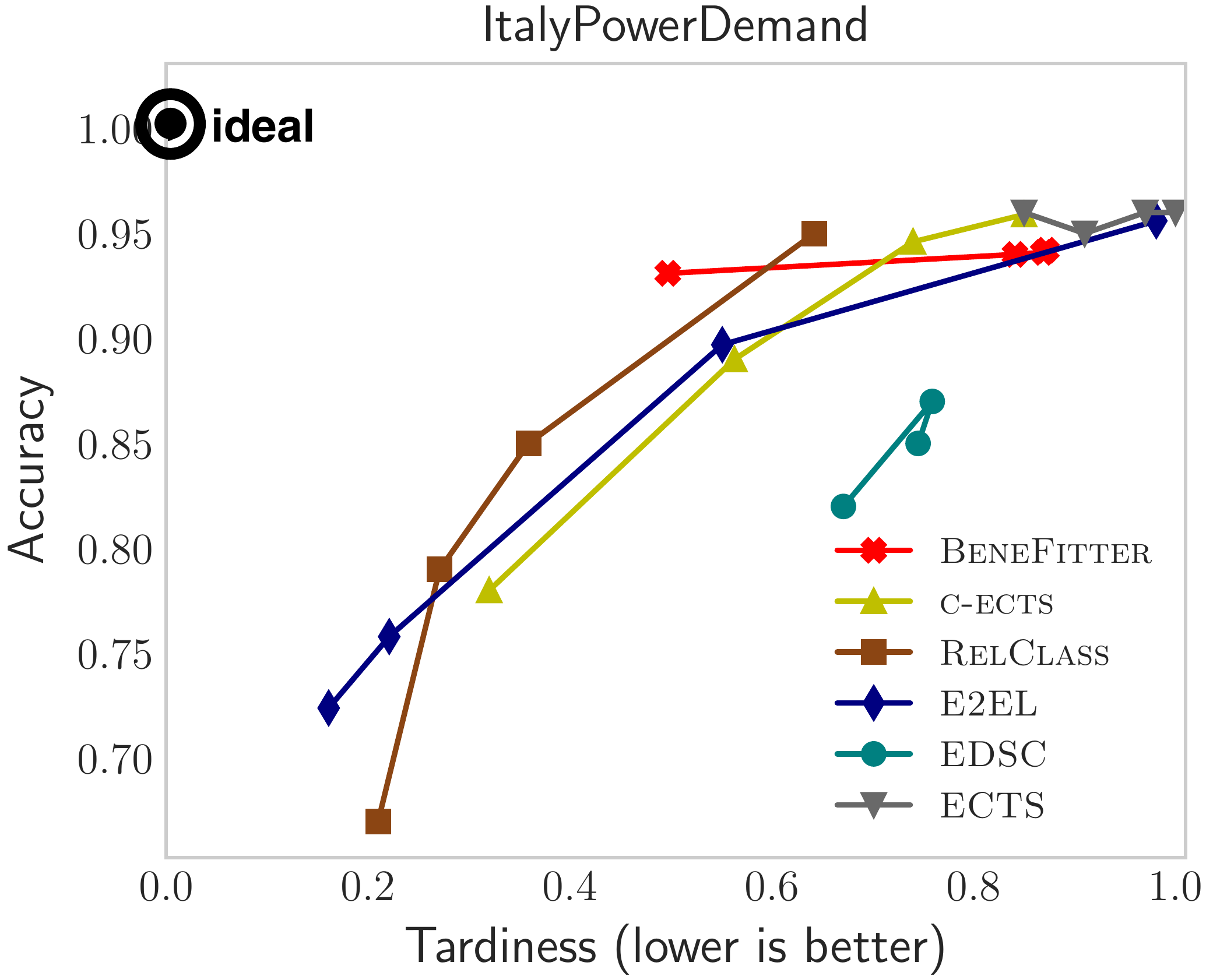}
	\end{subfigure}
	\begin{subfigure}{0.32\textwidth}
		\centering
		\includegraphics[scale=0.27]{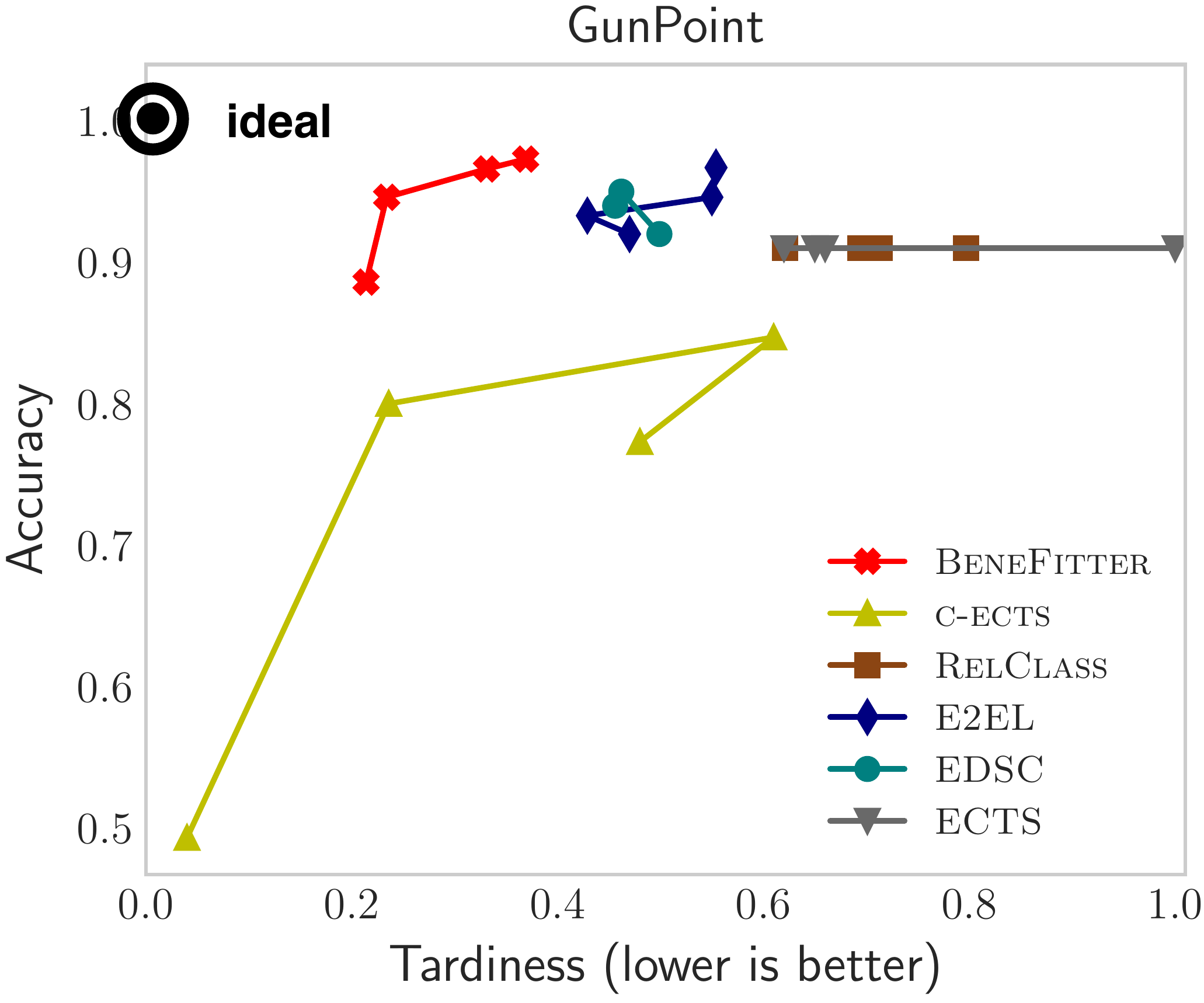}
	\end{subfigure}
	\begin{subfigure}{0.32\textwidth}
		\centering
		\includegraphics[scale=0.27]{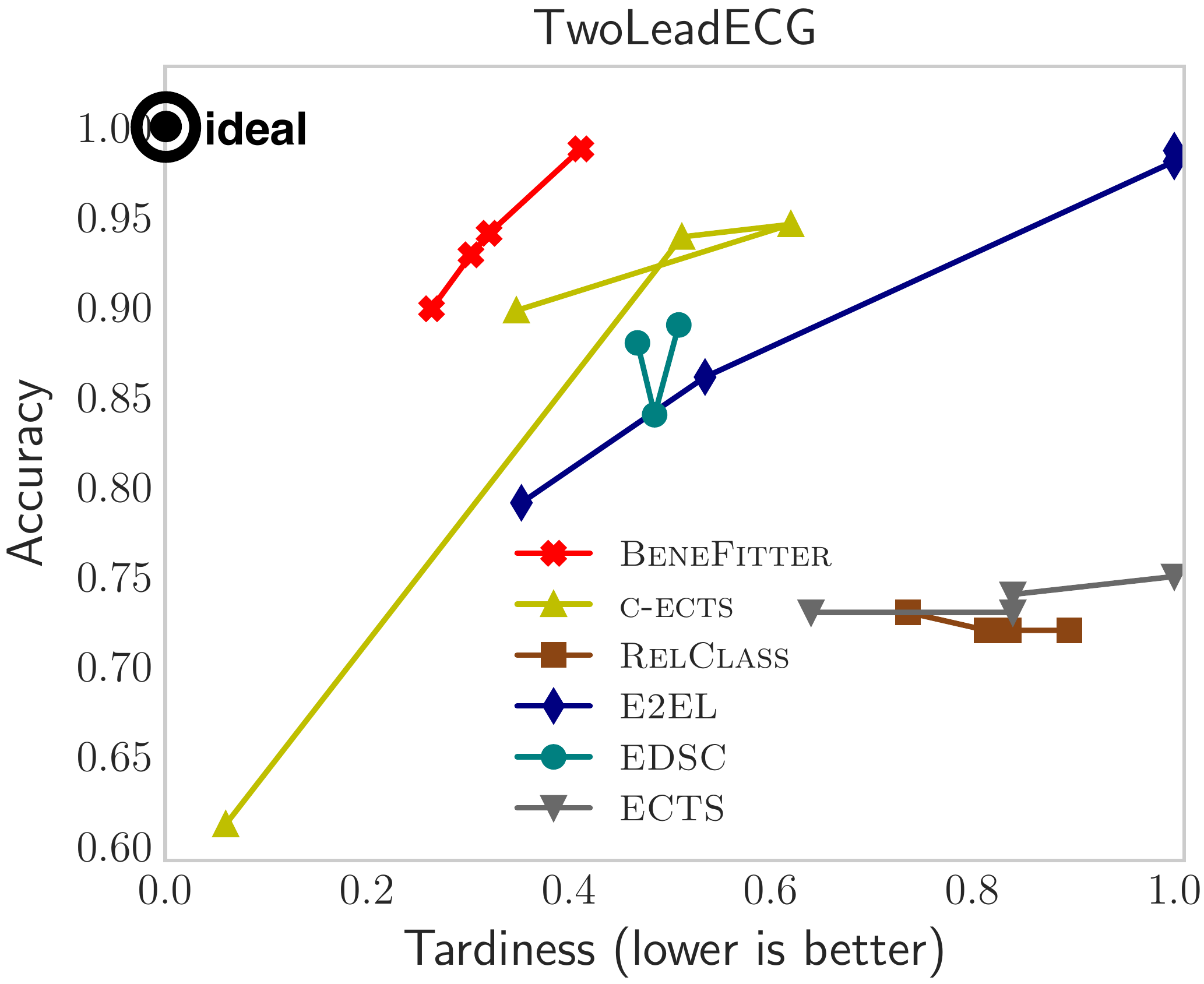}
	\end{subfigure}
	
	\begin{subfigure}{0.32\textwidth}
		\centering
		\includegraphics[scale=0.27]{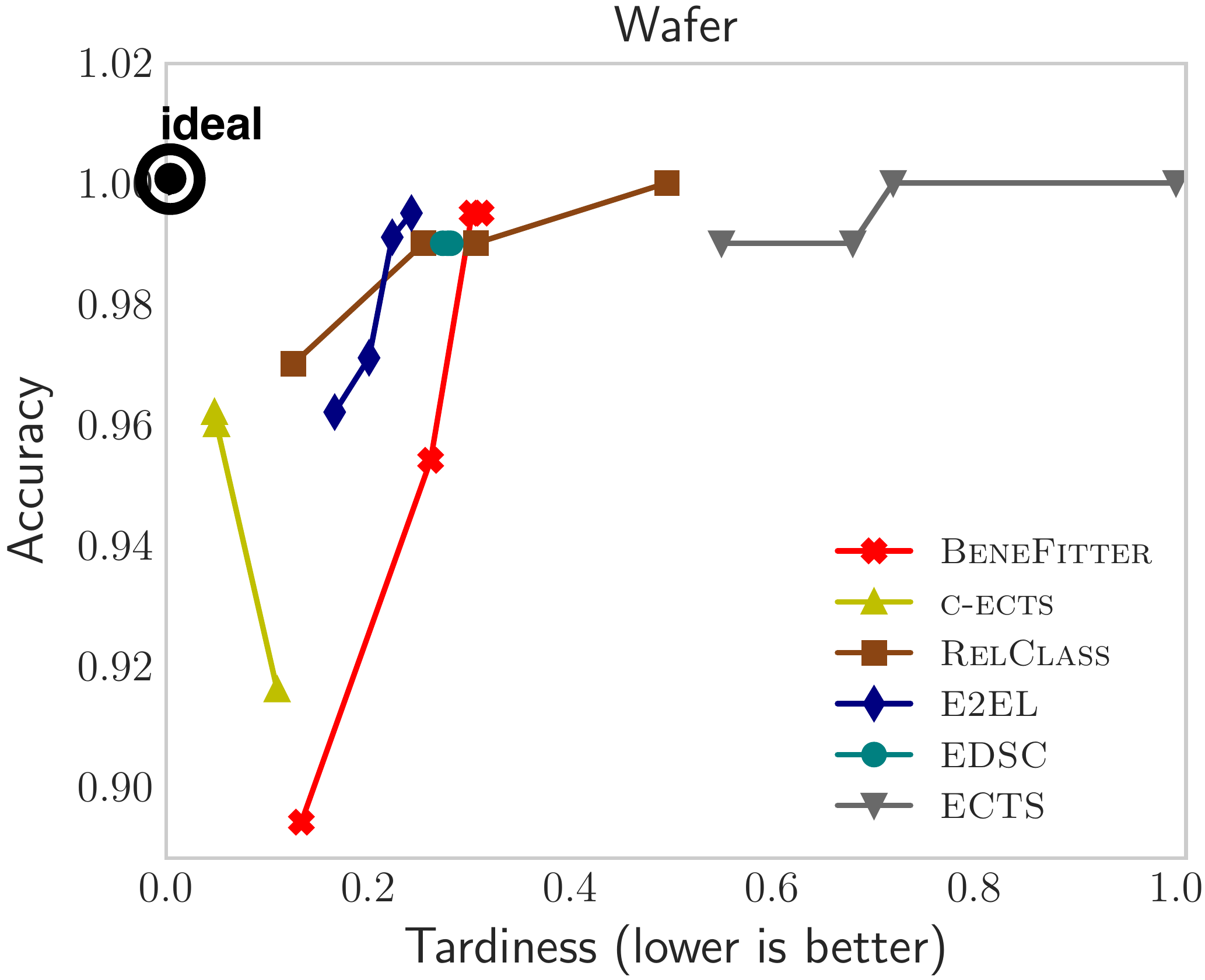}
	\end{subfigure}
	\begin{subfigure}{0.32\textwidth}
		\centering
		\includegraphics[scale=0.27]{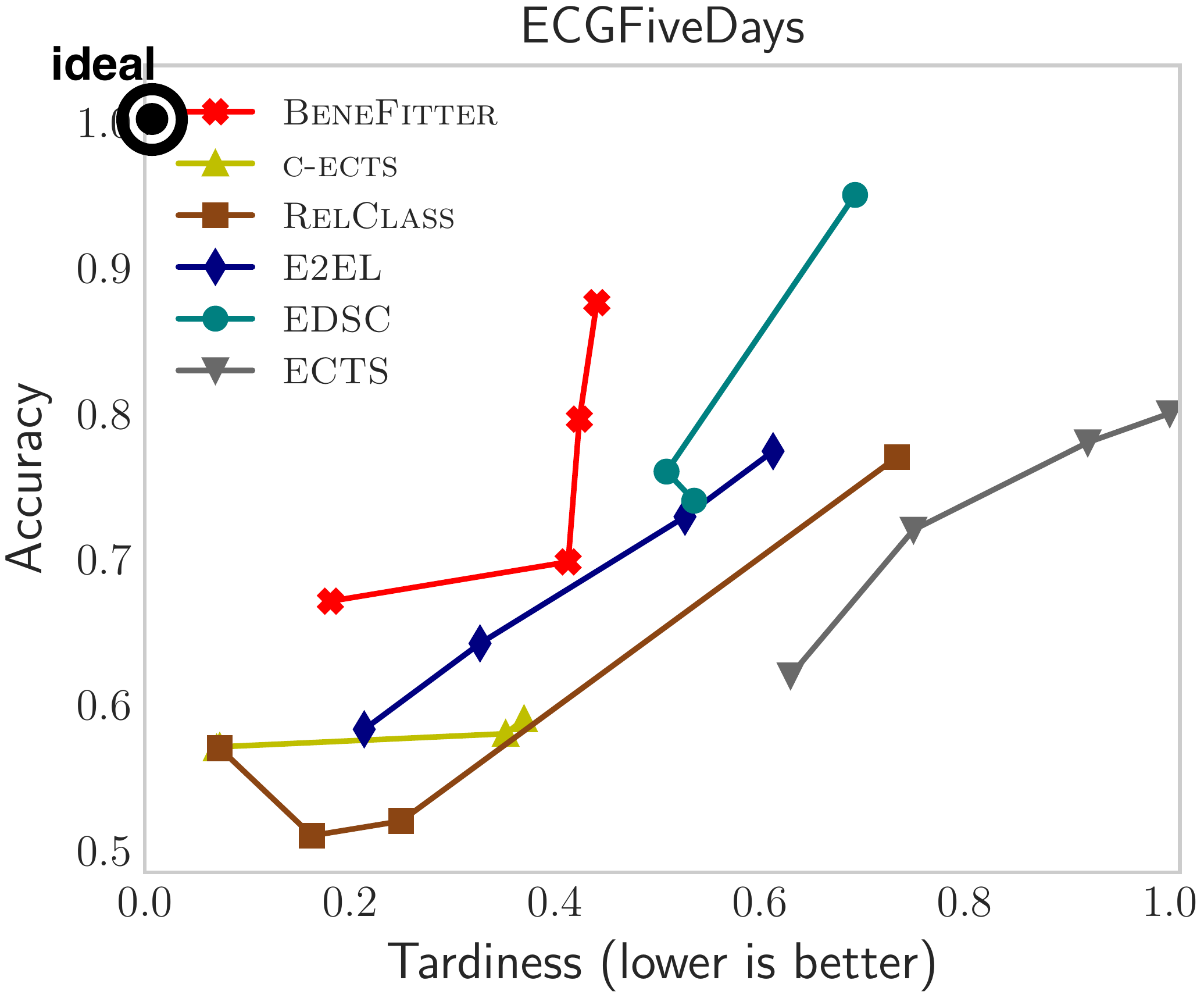}
	\end{subfigure}
	\begin{subfigure}{0.32\textwidth}
		\centering
		\includegraphics[scale=0.27]{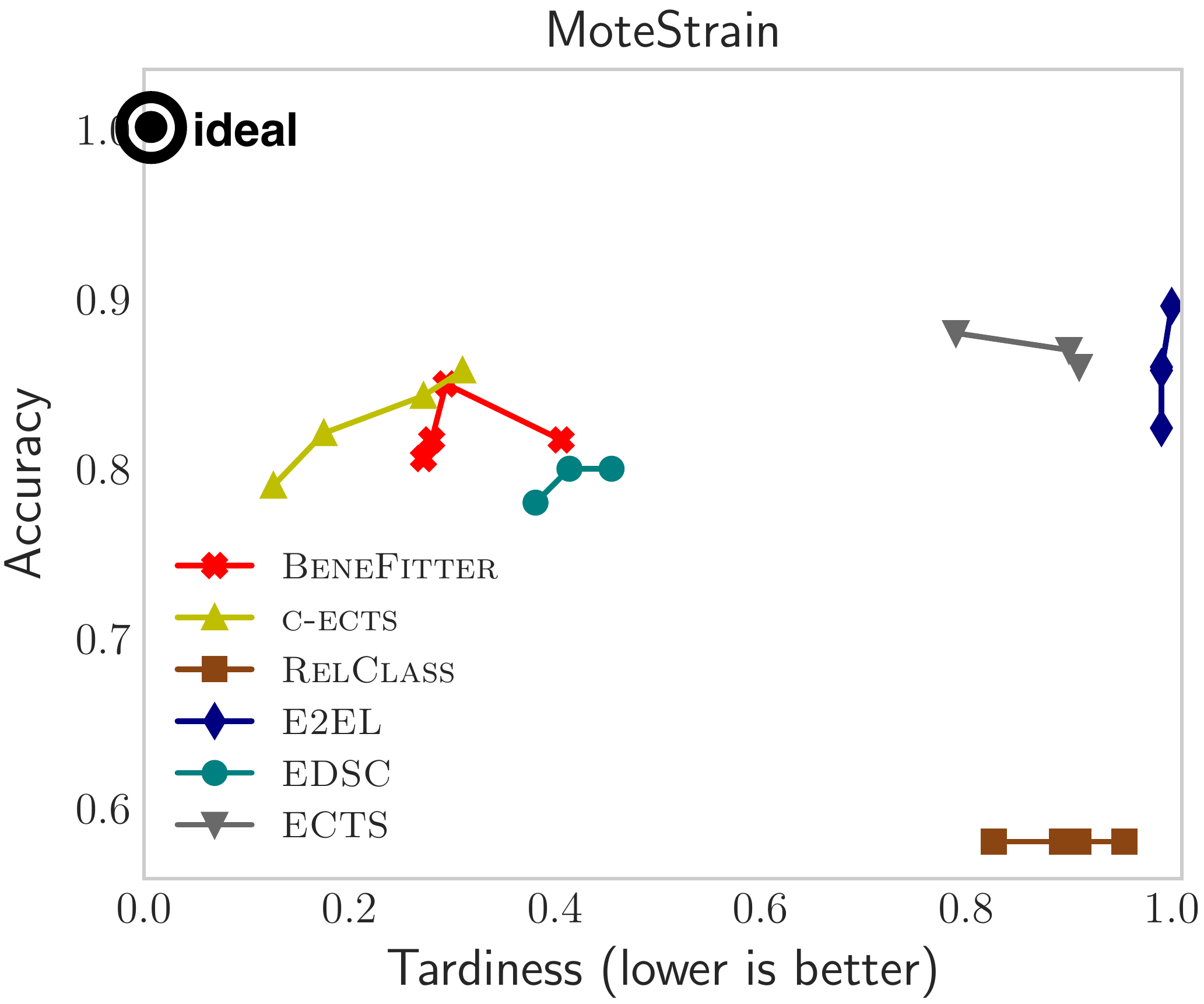}
	\end{subfigure}
	
	\begin{subfigure}{0.32\textwidth}
		\centering
		\includegraphics[scale=0.27]{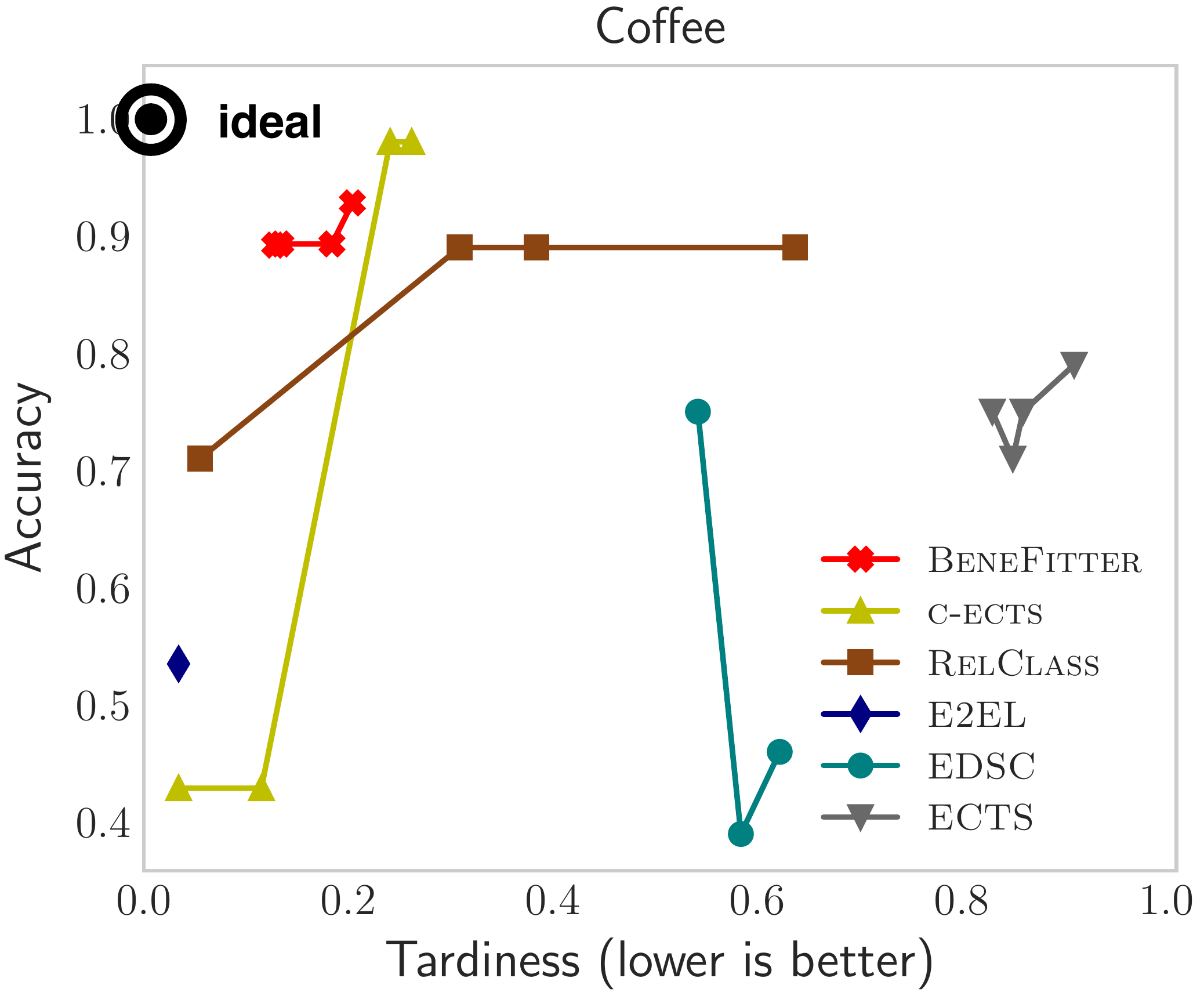}
	\end{subfigure}
	\begin{subfigure}{0.32\textwidth}
		\centering
		\includegraphics[scale=0.27]{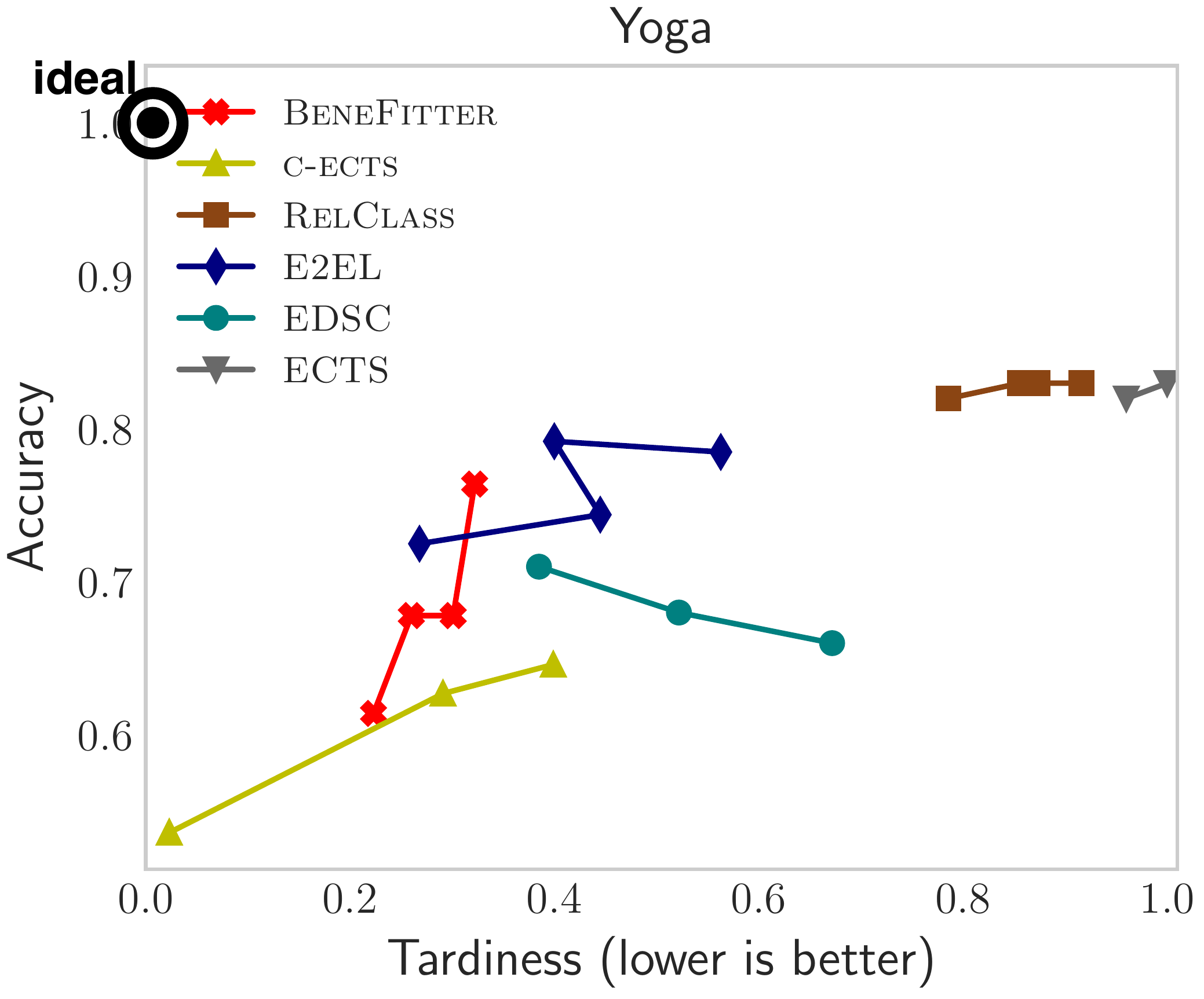}
	\end{subfigure}
	\begin{subfigure}{0.32\textwidth}
		\centering
		\includegraphics[scale=0.27]{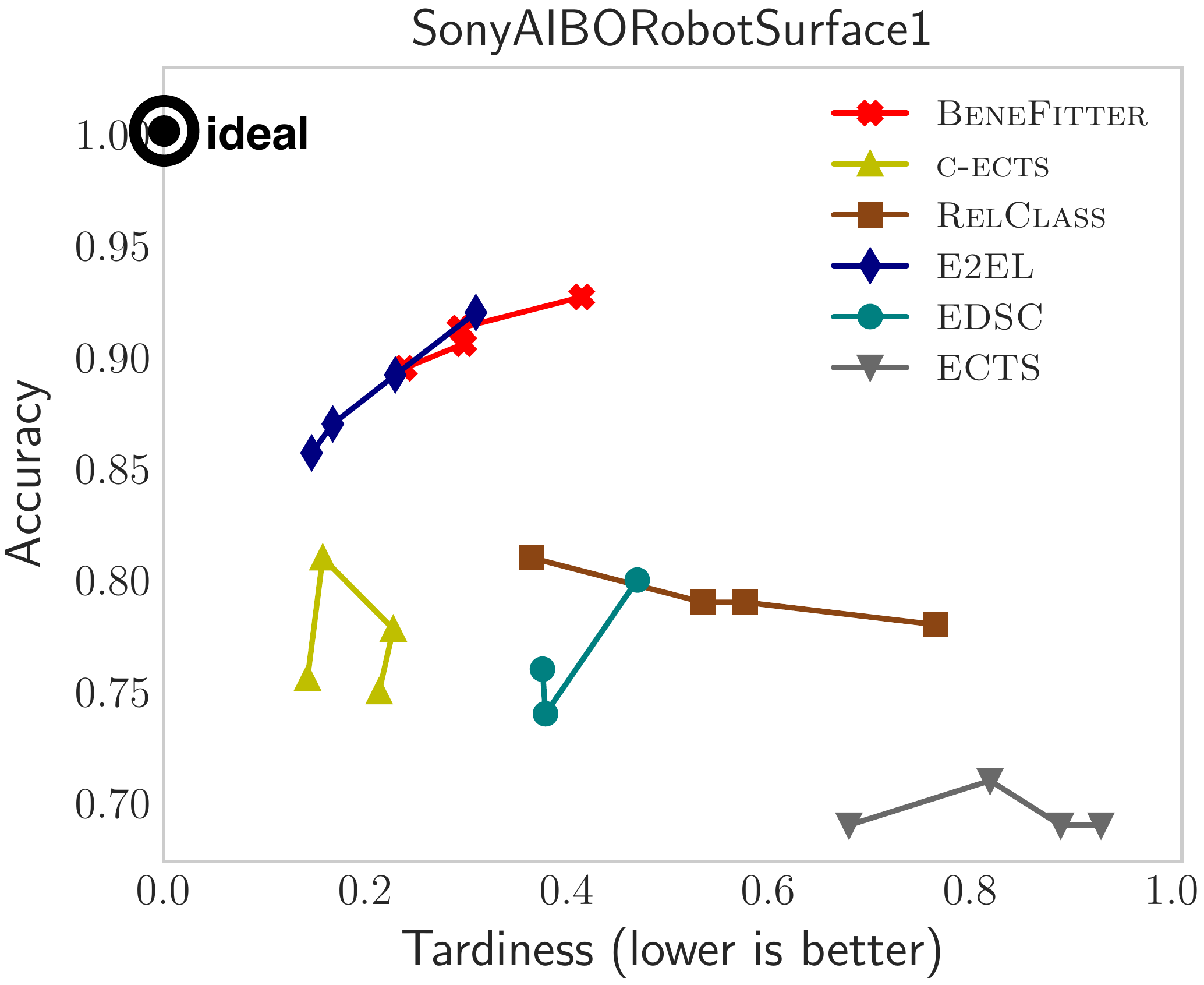}
	\end{subfigure}

	\caption{Comparison of methods based on accuracy versus tardiness trade-off for benchmark prediction tasks~(Sec.~\S\ref{subsec:evaluation}).\label{fig:pareto}}
\end{adjustbox}
\end{figure*}

\noindent{\bfseries Benchmark Prediction Tasks.~} To jointly evaluate the accuracy and earliness (tardiness -- lower is better), we plot accuracy against the tardiness to compare the Pareto frontier for each of the competing methods over 10 different benchmark datasets. In Fig.~\ref{fig:crown} and Fig.~\ref{fig:pareto}, we show the accuracy and tardiness trade-off for 10 benchmark UCR datasets. Each point on the plot represents the model evaluation for a choice of trade-off parameters reported in Table~\ref{tab:parameters}~(\S\ref{sec:supp_exp_details}). Note that \method dominates the frontiers of all the baselines in accuracy vs tardiness on five of the datasets. Moreover, our method appears on the Pareto frontier for four out of the remaining five for at least one set of parameters.

To further assess the methods, we report quantitative results in Table~\ref{tab:ucr_results} in terms of accuracy at a given tolerance of tardiness. We define an acceptable  level of tolerance $\in \{0.50, 0.75\}$ to indicate how much an application domain is indifferent to delay in decision up to the indicated level. For example a tolerance of $0.50$ indicates that the evaluation of the decisions is done at $t = 0.50\times L$, $L$ is the maximum length of sequence, and any decision made up to $t = 0.50\times L$ are considered for evaluation. In Fig.~\ref{fig:pareto}, we fix the x-axis at a particular tolerance and report the best accuracy to the left of the fixed tolerance in Table~\ref{tab:ucr_results}. The reported tolerance level indicates the average tolerance across the test time-series sequences. \method outperforms the competition seven times out of ten for a tolerance level $= 0.50$ indicating that our method achieves best performance using only the half of observations. The remaining three times our method is second best among all the competing methods. Similarly for tolerance $= 0.75$, \method is among the top two methods nine out of ten times.

\begin{table}[]
	\caption{ {\uline{\method wins}} most of the times. Accuracy on benchmark datasets against mean tardiness $\in \{0.5, 0.75\}$. \emph{Bold} represents best accuracy within a given tardiness tolerance, and the \emph{underline} represents the next best accuracy. `-' indicates that on average method requires more observations than the given tardiness tolerance. `\xmark' specifies non-applicability of a method on the dataset, and `{\sc dns}' shows that a method does not scale for the dataset.\label{tab:ucr_results}}
	\vspace{-0.1in}
	\resizebox{\columnwidth}{!}{
		\begin{tabular}{@{}rccccccc@{}}
			\toprule[1.2pt]
			 \multirow{3}{*}{\rotatebox{30}{Dataset}  }                             &   \multirow{3}{*}{\rotatebox{30}{Tardiness}} &   \multirow{3}{*}{\rotatebox{30}{ECTS} }&   \multirow{3}{*}{\rotatebox{30}{EDSC} }&  \multirow{3}{*}{ \rotatebox{30}{C-ECTS} }&   \multirow{3}{*}{\rotatebox{30}{RelClass} }&  \multirow{3}{*}{\rotatebox{30}{E2EL} }&   \multirow{3}{*}{\rotatebox{30}{\method} }\\
			 &      &                        &                   &                &                &         & 			 \\ 
			 &   ($\le$)   &                        &                   &                &                &         & 			 \\\midrule
				\multirow{2}{*}{ \rotatebox{30}{ECG200}}               & 0.50      & -                        & 0.84                     & 0.83                       & \gray{\ul 0.88}                   & 0.87                     & \blue{{\textbf{0.91}}}                  \\
			& 0.75      & -                        & 0.84                     & 0.83                       & \gray{\ul 0.89}                   & 0.87                     & \blue{\textbf{0.91}}                  \\\\
			
			\multirow{2}{*}{ \rotatebox{30}{ItalyPower}}
			& 0.50      & -                        & -                        & 0.78                       & 0.85                         & \gray{\ul 0.89}               & \blue{\textbf{0.93} }                 \\
			& 0.75      & -                        & 0.85                     & \gray{\ul 0.94}                 & \blue{\textbf{0.95}}                & 0.89                     & 0.93                           \\\\
			\multirow{2}{*}{ \rotatebox{30}{GunPoint}}
			& 0.50      & -                        & \gray{\ul 0.95}               & 0.80                       & -                            & 0.93                     & \blue{\textbf{0.97} }                 \\
			& 0.75      & 0.91                     & 0.95                     & 0.84                       & 0.91                         & \gray{\ul 0.96}               & \blue{\textbf{0.97}}                  \\\\
			
			\multirow{2}{*}{ \rotatebox{30}{TwoLeadECG}}
			& 0.50      & -                        & 0.88                     & \gray{\ul 0.89}                 & -                            & 0.79                     & \blue{\textbf{0.98}}                  \\
			& 0.75      & 0.73                     & 0.89                     & \gray{\ul 0.94}                 & 0.73                         & 0.86                     & \blue{\textbf{0.98}}                  \\\\
			
			\multirow{2}{*}{ \rotatebox{30}{Wafer}}     
			& 0.50      & -                        & \gray{\ul 0.99}               & 0.96                       & \blue{\textbf{1.0}}                 & \gray{\ul 0.99}               & \gray{\ul 0.99}                     \\
			& 0.75      & \textbf{1.0}             & \gray{\ul 0.99}               & 0.96                       & \blue{\textbf{1.0}}                 & \gray{\ul 0.99}               & \gray{\ul 0.99}                     \\\\
			
			\multirow{2}{*}{ \rotatebox{30}{ECGFiveDays}} 
			& 0.50      & -                        & -                        & 0.59                       & 0.57                         & \gray{\ul 0.64}               & \blue{\textbf{0.87}}                  \\
			& 0.75      & 0.72                     & \blue{\textbf{0.95}}            & 0.59                       & 0.77                         & 0.77                     & \gray{\ul 0.87}                     \\\\
			
			\multirow{2}{*}{ \rotatebox{30}{MoteStrain}}        
			& 0.50      & -                        &\gray {\ul 0.8}                & \blue{\textbf{0.85} }             & -                            & -                        & \blue{\textbf{0.85} }                 \\
			& 0.75      & -                        & \gray{\ul 0.8}                & \blue{\textbf{0.85}  }            & -                            & -                        & \blue{\textbf{0.85} }                 \\\\
			
			\multirow{2}{*}{ \rotatebox{30}{Coffee}} 
			& 0.50      & -                        & -                        & \blue{\textbf{0.98}}              & 0.89                         & 0.53                     &\gray {\ul 0.93}                     \\
			& 0.75      & -                        & 0.75                     & \blue{\textbf{0.98} }             & 0.89                         & 0.53                     & \gray{\ul 0.93}                     \\\\
			
			\multirow{2}{*}{ \rotatebox{30}{Yoga}}               
			& 0.50      & -                        & 0.71                     & 0.64                       & -                            & \blue{\textbf{0.79}  }          & \gray{\ul 0.76}                     \\
			& 0.75      & -                        & 0.71                     & 0.64                       & -                            & \blue{\textbf{0.79} }           & \gray{\ul 0.76}                     \\\\
			
			\multirow{2}{*}{ \rotatebox{30}{SonyAIBO}} 
			& 0.50      & -                        & 0.80                     & \gray{\ul 0.81}                 & \gray{\ul 0.81}                   & \blue{ \textbf{0.92}}           & \blue{\textbf{0.92}  }                \\
			& 0.75      & 0.69                     & 0.80                     & \gray{\ul 0.81}                 & \gray{\ul 0.81}                   & \blue{ \textbf{0.92}}            &\blue{ \textbf{0.92}  }                \\\\ \midrule \\
			
				\multirow{1}{*}{ {Endomondo}} 
			& 0.50      & \xmark                        &      {\sc dns}                   & \xmark                  & \xmark                         & \blue{\textbf{0.68}}  & \gray{\ul 0.66}                 \\ \bottomrule
		\end{tabular}
	}
\vspace{-0.2in}
\end{table}

\noindent{\bfseries Endomondo Activity Prediction.~} We run the experiments on full Endomondo dataset (a large scale dataset) to compare \method with baseline {\scshape{E2EL}} (other baselines do not scale) for one set of earliness-accuracy trade-off parameters. 
We, first, compare the two methods on a sampled dataset -- with $1000$ time series instances -- evaluated for a choice of trade-off parameters 
(see Fig.~\ref{fig:endomondo} in Appendix~\ref{app:results}).
We select the parameters that yields a performance closest to \emph{ideal} indicated.
With the selected parameters, comparison of two methods on large-scale Endomondo activity prediction dataset are reported in Table~\ref{tab:ucr_results} (last row). 
We report the accuracy of the two methods by fixing their tardiness at $\le 0.5$. Notice that the two methods are comparable in terms of the prediction performance while using less than half the length of a sequence for outputting a decision.\looseness=-1

The quantitative results suggest a way to choose the best classifier for a specified tolerance level for an application. In critical domains such as medical, or predictive maintenance a lower tolerance would be preferred to save cost. In such domains, \method provides a clear choice for early decision making based on the benchmark dataset evaluation.

\subsubsection*{\bfseries [Q2] Efficiency}
\begin{figure}[!h]
	\centering
	\begin{tabular}{cc}
		\includegraphics[width=0.22\textwidth]{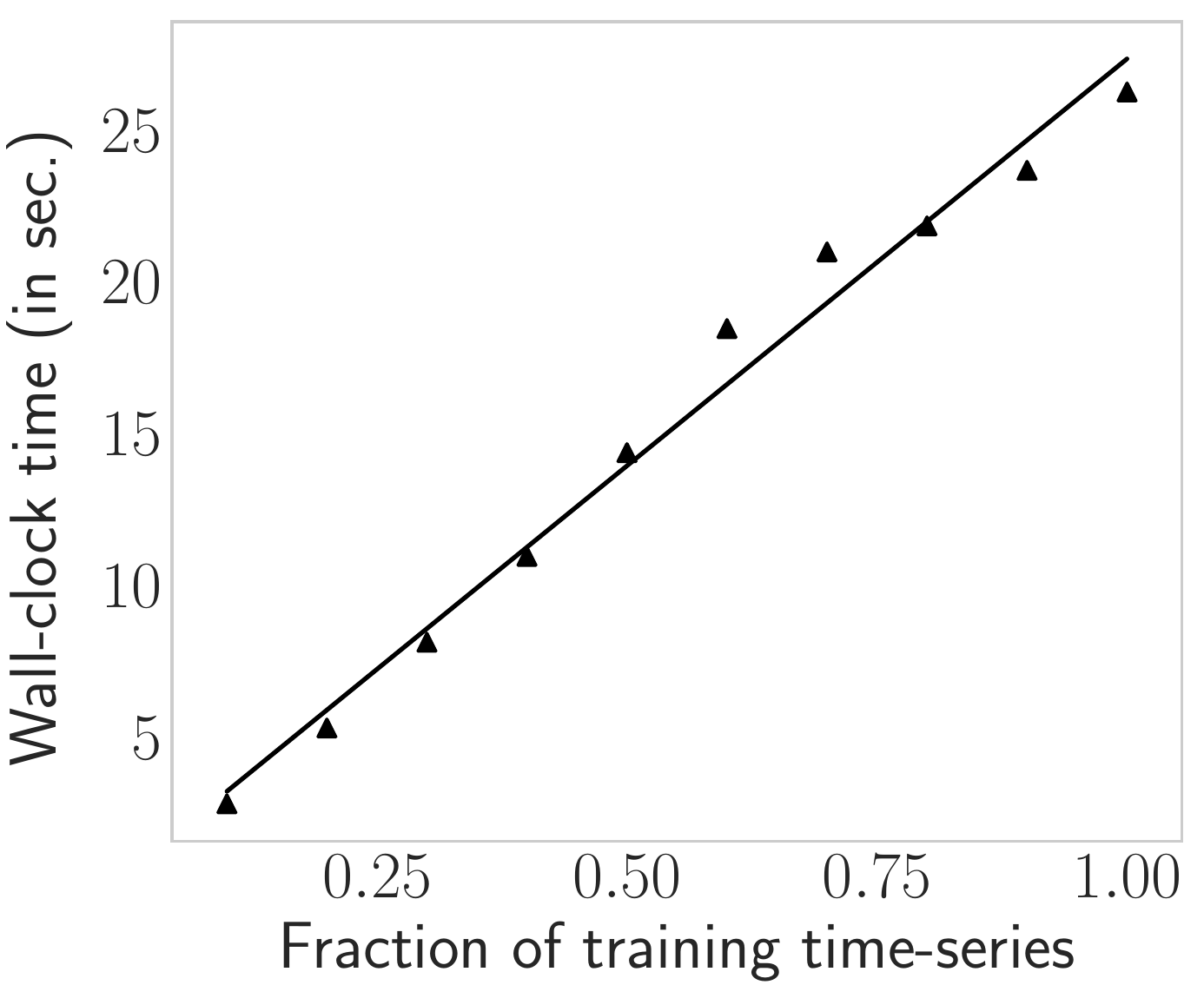} &
		\includegraphics[width=0.23\textwidth]{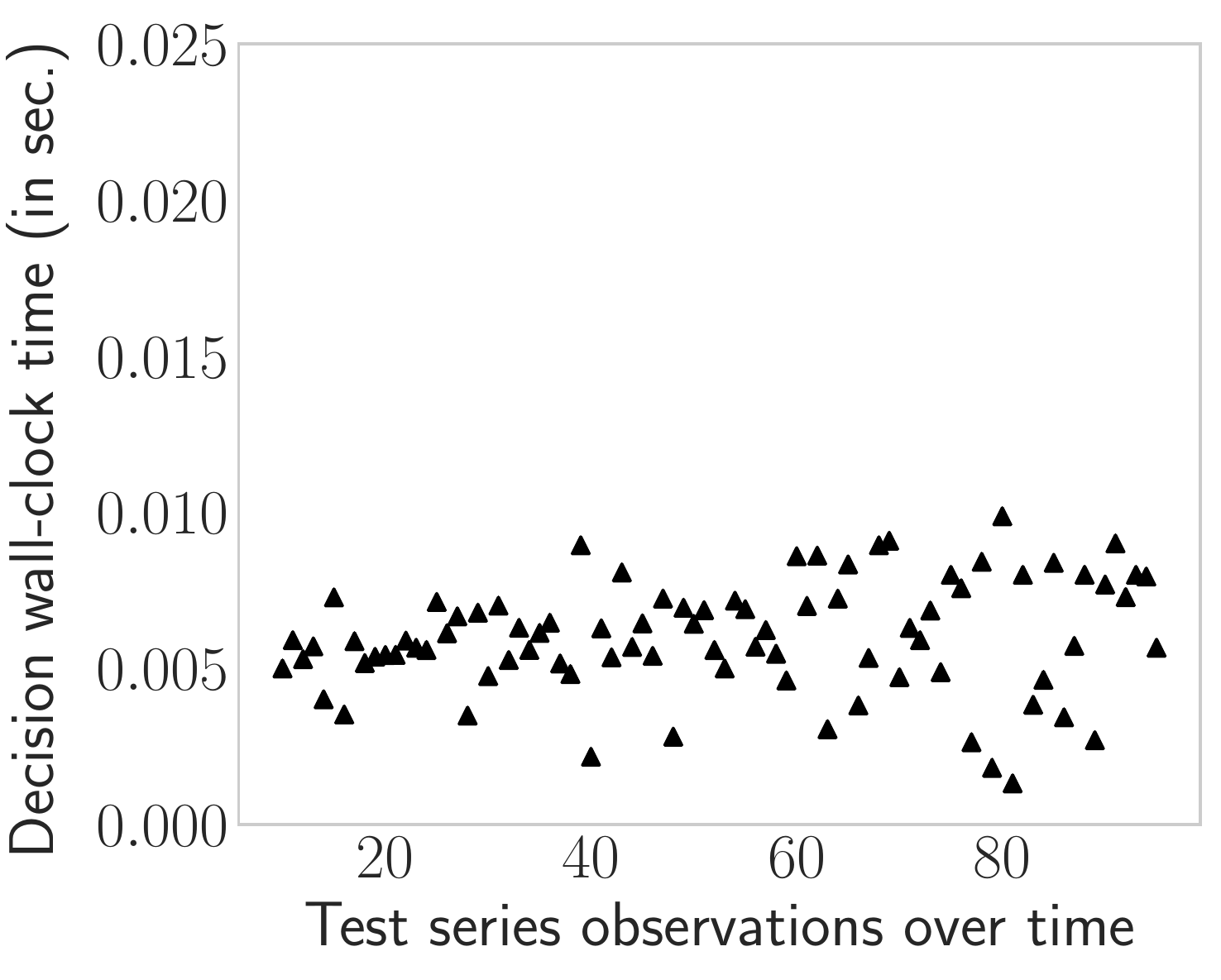}
	\end{tabular}
	\vspace{-0.15in}
	\caption{(left) \method scales linearly with the number of time-series, and (right) provides constant-time decision. \label{fig:runtime}}
		\vspace{-0.1in}
\end{figure}

Fig.~\ref{fig:runtime} shows the scalability of \method with the number of training time-series and number of observations per time series at test time. We use ECG200 dataset from UCR benchmark to report results on runtime.

\noindent{\bfseries {{Linear training time}:~}} We create ten datasets from the ECG200 dataset by retaining a fraction $\in \{0.1, 0.2, \dots, 1.0\}$ of total number of training instances. For a fixed set of parameters, we train our model individually for each of the created datasets. The wall-clock running time is reported against the fraction of training sequences in Fig.~\ref{fig:runtime} (left). The points in the plot align in a straight line indicating that \method scales linearly with number of sequences.

\noindent{\bfseries {Constant test time}:~}  We now evaluate \method runtime by varying the number of observations over time. For this experiment, we retain the hidden state of an input test sequence up to time $(t-1)$. When a new observation at time $t$ arrives, we update the hidden state of the RNN cell using the new observation and compute the predicted benefit based on updated state.   Fig.~\ref{fig:runtime} (right) plots the wall-clock time against each new observation. The time is averaged over test set examples. The plot indicates that we get real-time decision in constant time.

The efficiency of our model makes it suitable for deployment for real time prediction tasks.
\subsubsection*{\bfseries [Q3] Discoveries} 
In this section, we present an analysis of \method highlighting some of the salient aspects of our proposed framework on ICU patient outcome task. In particular, we discuss how our method explains the \benefit{} prediction by highlighting the parts of inputs that contributed most for the prediction, and how our \benefit{} formulation assists with model evaluation.

\begin{figure}[!t]
	\centering
	\includegraphics[width=\linewidth, height=0.23\textwidth]{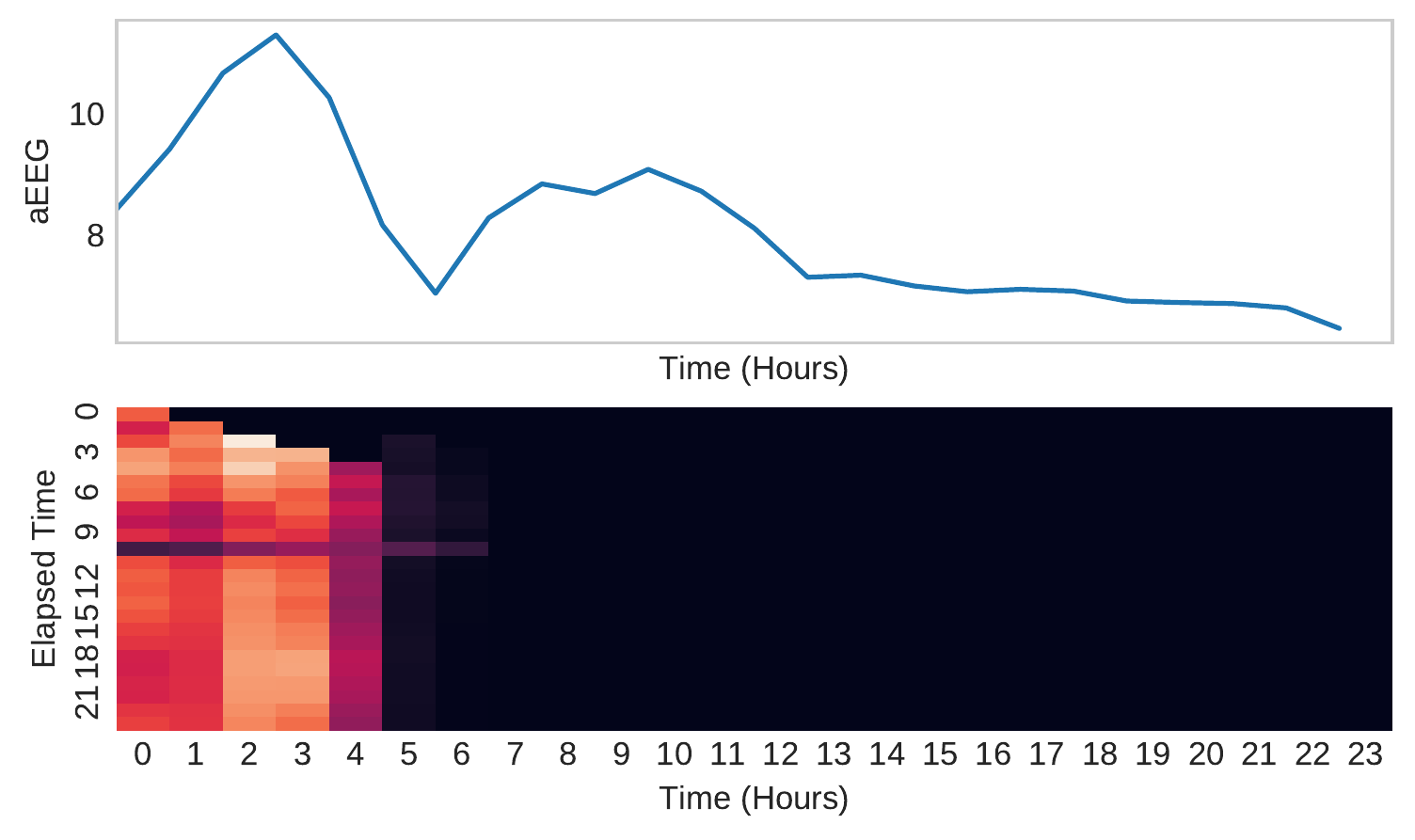}
	\vspace{-0.3in}
	\caption{Input test sequence with corresponding attention weights evaluated at $t = {0,\ldots,23}$. Model decision at $t=4$. \label{fig:attn}}
	\vspace{-0.1in}
\end{figure}

\noindent {\bfseries Explaining Benefit Estimation} Our method utilizes the attention mechanism (see \S\ref{sec:meth}) in the RNN network for {\tt benefit} regression. The model calculates \emph{weights} corresponding to each hidden state $\bh_t$. These weights can indicate which of the time dimensions model focuses on to estimate the benefit for the current input series. In Fig.~\ref{fig:attn}, we plot one dimension of the input time series from EEG dataset. This dimension corresponds to amplitude of the EEG (aEEG) when measured in left hemisphere of the brain. The input sequence is taken from the hourly sampled dataset. Note that there are sharp rise and fall in the aEEG signal from $t=1$ to $t=5$, and  from $t=5$ to $t=13$. We input the $107$ dimensional sequence to \method along with the aEEG signal. The model outputs the attention weights corresponding to each time-dimension of the inputs shown in Fig.~\ref{fig:attn} as a heatmap~(dark colors indicate lower weights, lighter colors indicate higher weights). \method outputs a decision at $t=4$, however we evaluate the model at further time steps. Note that the each row of heat map represents evaluation of input at $t=0, \dots, 23$. For each evaluation, we obtain a weight signifying the importance of a time dimension which are plotted as heatmap. X-axis of heat map corresponds to time dimension, and y-axis of the heatmap corresponds to the evaluation time  step of the input. Observe that the attention places higher weights towards the beginning of the time series where we observe the crests and troughs of the signal. 
The visualization of importance weights is useful in critical applications where each decision involves a high cost. The estimated benefit along with the importance information 
can assist a clinician with the decision making.

\noindent {\bfseries Assisting with Model Evaluation} In the clinical setting with comatose patients, there is a natural \textit{cost} associated with an inaccurate prediction and \textit{savings} obtained from knowing the labels early. The {\tt benefit} modeling captures the overall value of outputting a decision. Though, we use this value for learning a regression model, the {\tt benefit} formulation could be used as an evaluation metric to asses the quality of a predictive model.
If we know the domain specific unit-time savings $s$ and misclassification cost $M$, we can then evaluate a model performance for that particular value of $s$ and $M$. Table~\ref{tab:discovery} reports evaluation of \method for various values of $M/s$ with model trained using the same values of $M/s$. We notice that with increasing $M/s$ we improve the precision of the model, however increased $M/s$ also results in higher penalty for any misclassification. For our model trained on hourly sampled EEG data, we observe that values above $M/s =300$, results in overall negative {\tt benefit} averaged over test data. Assuming unit-time savings $s=\$4000$, we can tolerate lawsuit costs up to $\$1.2$million for $M/s=300$. Similarly, any model can be evaluated to assess its usefulness using our {\tt benefit} formulation as an evaluation measure in critical domains.  

\begin{table}[!ht]
	\caption{Training and evaluation of \method for $M/s=\{100, 200, 300, 400\}$ on EEG hourly sampled data.\label{tab:discovery}}
	\vspace{-0.1in}
	\resizebox{0.96\columnwidth}{!}{
	\begin{tabular}{@{}rcccccr@{}}
		\toprule
		$M/s$   & Precision & Recall & F1 score & Accuracy & Tardiness & Benefit \\ \midrule
		100 & 0.80      & 0.68   & 0.73     & 0.83     & 0.64      & 2737    \\
		200 & 0.80      & 0.67   & 0.71     & 0.82     & 0.68      & 1032    \\
		300 & 0.82      & 0.67   & 0.74     & 0.84     & 0.68      & 156     \\
		400 & 0.82      & 0.69   & 0.75     & 0.84     & 0.70      & -1326   \\ \bottomrule
	\end{tabular}}
	\vspace{-0.1in}
\end{table}

\section{Conclusions}
\label{sec:concl}

In this paper, we consider the benefit-aware early prediction of health outcomes for ICU patients and proposed \method that is designed to effectively handle multi-variate and variable-length signals such as EEG recordings.
We made multiple contributions.
\begin{enumerate}
	\item {\bf Novel, cost-aware problem formulation}: \method infuses the incurred savings from an early prediction as well as the cost from misclassification into a unified target called \benefit. Unifying these two quantities allows us to directly estimate a \textit{single} target, i.e., {\tt benefit}, and importantly 
	dictates \method exactly \textit{when} to output a prediction: whenever estimated {\tt benefit} becomes positive.
	
	\item {\bf Efficiency and speed}: The training time for \method is linear in the number of input sequences, and it can operate under a streaming setting to update its decision
	based on incoming observations.
	
	\item {\bf Multi-variate and multi-length time-series}: \method is designed to handle multiple time sequences, of varying length, suitable for various domains including health care.
	
	\item {\bf Effectiveness on real-world data}: We applied \method in early prediction of health outcomes on ICU-EEG data where \method provides 
	up to $2\times$ time-savings as compared to competitors while achieving equal or better accuracy. \method also outperformed or tied with  top competitors on other real-world benchmarks.

\end{enumerate}

\bibliographystyle{ACM-Reference-Format}
\bibliography{IEEEabrv,BIB/reference.bib,BIB/dhivya}

\clearpage
\appendix

\section{Appendix}
\label{sec:supplement}
In this section, we first give a detailed description of the datasets used in the experiments. We then provide model training details for \method, and present the hyper-parameter configurations used in \method as well as the competing methods for the benchmark experiments.
\setcounter{table}{0}
\renewcommand{\thetable}{A\arabic{table}}

\subsection{Dataset Description}
\label{sec:supp_data_desc}
\noindent \textbullet $\;$   {\bfseries Benchmark Datasets.~} Our benchmark datasets consist of 10 two-class time-series classification datasets from the UCR repository~\cite{UCRArchive}. The datasets cover diverse domains and have diverse range of series length. The UCR archive provides the train/test split for each of these datasets, which we retain in our experiments. 

\noindent\textbullet $\;$  {\bfseries Endomondo Dataset.~} Endomondo is a social fitness app that tracks numerous fitness attributes of the users. We use the web-scale Endomondo dataset~\cite{ni2019modeling}~(See Table~\ref{tab:datadesc}) for the early activity prediction task. The data includes various measurements such as heart rate, altitude and speed, along with contextual data such as user id and activity type. For the task of early activity prediction, we use heart rate and altitude signals for early prediction of the type of activity, specifically biking vs. running. (Note that we leave out signals like speed and its derivatives which make the classification task too easy.) 

\subsection{Experimental Settings}
\label{sec:supp_exp_details}
\subsubsection{Model Training Details}
We define the outcome prediction problem as a regression task on the {\tt benefit}, as presented in \S\ref{sec:meth}. The training examples represent the sequences observed up to time $t$ along with their corresponding expected {\tt benefit} at time $t$. We then split the training examples to use $90\%$ of the sequences for training the RNN model and remaining $10\%$ for validation. We select our model parameters based on the evaluation on validation set. We have two sets of hyper-parameters: one corresponding to our {\tt benefit} formulation that are $s$ and $M$, and the other for the RNN model. The hyper-parameter grid for the RNN model includes 
the dimension of the hidden representation $\in \{16, 32\}$ and the learning rate $\eta \in \{0.01, 0.001\}$. 
For \method, we fix $s$ at 1 and vary $M/s$ for model selection. The hyper-parameter grid for our {\tt benefit} formulation is $M/s \in \{0.5, 1.0, 1.5, 2.0\} \times \max(\{L_i\}_{i=1}^n)$, where $L_i$ is the length of training series $i$. For the general multi-class problem we tune for an additional hyper-parameter $\Delta$ to predict the class label. We set $\Delta \in \{0.4, 0.5, 0.6, 0.7\}$ of the maximum difference between the expected {\tt benefit} for the two-class labels for a given training series. \method outputs a decision when the predicted {\tt benefit} is positive and at least $\Delta$ higher compared to the predicted {\tt benefit} of other labels.

For the learning task, we use the mean squared error loss function and Adam optimizer~\cite{kingma2014adam} for learning the parameters. 
 The loss corresponding to class $l$ is given by
\begin{align*}
\mathcal{J} = \frac{1}{|N_{train}|} \sum_{b_{l} \in \mathcal{B}_l} \sum_{t=1}^{L_i} (\hat{b}_{itl} - b_{itl} )^2
\end{align*}
where $|N_{train|}$ is the number of total count of time steps for all the sequences in the training set, $L_i$ is the length of each sequence, and $\mathcal{B}_l$ denotes the expected {\tt benefit} for the class $l$. 
We use Keras
and Pytorch to implement our models.

\subsubsection{Benchmark Experiments -- Hyper-parameters}
We compared the performance of \method against six competing methods on benchmark datasets. In Table~\ref{tab:parameters}, we report the different hyper-parameter configurations for each method that provides a trade-off between accuracy and earliness.
\begin{table}[H]
	\caption{Earliness and accuracy trade-off parameters for each of the methods.\label{tab:parameters}}
	\vspace{-0.1in}
	\begin{tabular}{@{}rl@{}}
		\toprule
		Method                 & Model Training Hyper-parameters                                                          \\
		\midrule
		ECTS                   & {support} $\in \{0.1, 0.2, 0.4, 0.8 \}$       \\
		EDSC                   & {Chebyshev parameter} $\in \{2.5, 3.0, 3.5\}$ \\
		C-ECTS                 &  {delay cost} $\in \{0.0005,0.001,0.005,0.01\}$ \\
		RelClass               &   {reliability} $\in {0.001, 0.1, 0.5, 0.9}$  \\
		E2EL                   &  {earliness trade-off}  $\alpha \in \{0.6, 0.7, 0.8, 0.9\}$ \\
		\method &  $M/s \in \{0.5, 1.0, 1.5, 2.0\} \times \max(\text{series length})$ \\
		\bottomrule
	\end{tabular}
\end{table}

\subsection{Results}
\label{app:results}
\begin{figure}[H]
	\centering
	\includegraphics[scale=0.32]{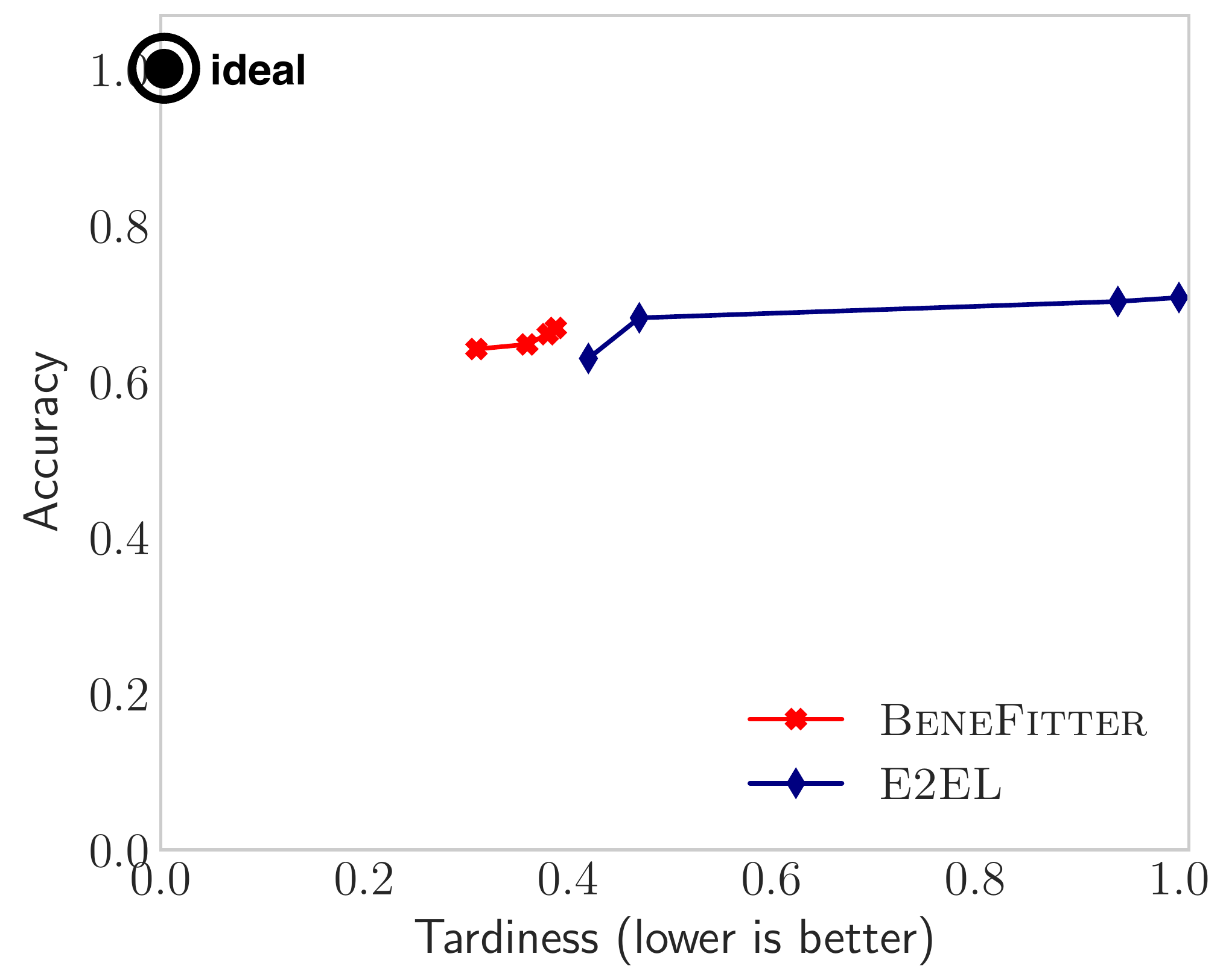}
	\caption{Accuracy versus tardiness for sampled Endomondo.\label{fig:endomondo}}
\end{figure}

\end{document}